\documentclass[journal]{IEEEtran}
\usepackage{cite}
\ifCLASSINFOpdf
\else
\fi
\usepackage{graphicx}
\usepackage{amsmath,amsthm,amssymb}
\usepackage{multirow}
\usepackage{color}
\usepackage{subfig}
\usepackage{soul}
\usepackage{breqn}
\usepackage[table,xcdraw]{xcolor}
\usepackage{stackengine}
\usepackage{tabularx}
\usepackage{todonotes}
\usepackage{booktabs}
\usepackage{siunitx}
\usepackage{enumerate}
\usepackage{hyperref}
\usepackage{extarrows}
\usepackage{soul}
\usepackage{todonotes}

\begin{document}
% Titles are generally capitalized except for words such as a, an, and, as, at, but, by, for, in, nor, of, on, or, the, to and up, which are usually not capitalized unless they are the first or last word of the title.

%I have one comment: we need to modify the title because adversarial attacks are against trained models not input data nor their representations; in fact classifiers are victims of the adversary. I would recommend something like these: 

%Robustness of Sound Classifiers Against Adversarial Attacks from 2D Representation Perspective

%Characterizing the Dependency of Adversarial Attacks to 2D Sound Representations

% original
%\title{Multi-Discriminator Sobolev GAN for Defensing Against End-To-End Speech Adversarial Attacks}

% too long
\title{RCC-GAN: Regularized Compound Conditional GAN for Large-Scale Tabular Data Synthesis}

% Short but Speech is not present
%\title{Multi-Discriminator Sobolev GAN for Defensing Against Adversarial Attacks}

%\title{Threat of Adversarial Attacks for 1D Audio Signals and their 2D Representations}

%\title{A Robust Approach for Securing Audio Systems Against Adversarial Attacks}

%\title{Robustness of \ak{2D Audio Representations} Against Adversarial Attacks}

%\title{Robustness of Sound Classifiers Against Adversarial Attacks from 2D Representation Perspective}

%\title{Comparison of Different Sound Representations Against Adversarial Attacks}


\author{Mohammad Esmaeilpour$^{\mathsection \dagger}$,~\IEEEmembership{Member,~IEEE,}
 Nourhene Chaalia$^\dagger$, Adel~Abusitta$^\ddagger$,~\IEEEmembership{Senior Member,~IEEE,}
 François-Xavier~Devailly$^\dagger$, Wissem Maazoun$^\dagger$,
 Patrick~Cardinal$^\mathsection$,~\IEEEmembership{Fellow Member,~IEEE}
\thanks{$^{\mathsection}$\'{E}cole de Technologie Sup\'{e}rieure (\'{E}TS), Universit\'{e} du Qu\'{e}bec, Montr\'{e}al, Qu\'{e}bec, Canada. $^{\dagger}$Fédération des Caisses Desjardins du Québec. $^{\ddagger}$University of Windsor, Ontario, Canada.
e-mail addresses: \\(1) mohammad.esmaeilpour.1@ens.etsmtl.ca,\\ (2) mohammad.esmaeilpour@desjardins.com.}% <-this % stops a space
%\thanks{Revised manuscript submitted August 19, 2019. Accepted for publication.}}
%\thanks{Accepted for publication November, 2019.}}
}

% ^------------^------------^----Do not want these spaces!

\markboth{Submitted to IEEE Transactions on Neural Networks and Learning Systems,~Vol.~X, No.~X, May~2022}%
{Esmaeilpour \MakeLowercase{\textit{et al.}}: IEEE Transactions on Information Forensics and Security}
% *** name in the headers of peer review papers. ***



\maketitle

\begin{abstract}
%%%%%%%%%%%%%%%%%%%%%%%%%%%%%%%%%%%%%%%%%%%%%%%%%%%%%%%%%%%%%%%%%%%%
This paper introduces a novel generative adversarial network (GAN) for synthesizing large-scale tabular databases which contain various features such as continuous, discrete, and binary. Technically, our GAN belongs to the category of class-conditioned generative models with a predefined conditional vector. However, we propose a new formulation for deriving such a vector incorporating both binary and discrete features simultaneously. We refer to this noble definition as compound conditional vector and employ it for training the generator network. The core architecture of this network is a three-layered deep residual neural network with skip connections. For improving the stability of such complex architecture, we present a regularization scheme towards limiting unprecedented variations on its weight vectors during training. This regularization approach is quite compatible with the nature of adversarial training and it is not computationally prohibitive in runtime. Furthermore, we constantly monitor the variation of the weight vectors for identifying any potential instabilities or irregularities to measure the strength of our proposed regularizer. Toward this end, we also develop a new metric for tracking sudden perturbation on the weight vectors using the singular value decomposition theory. Finally, we evaluate the performance of our proposed synthesis approach on six benchmarking tabular databases, namely Adult, Census, HCDR, Cabs, News, and King. The achieved results corroborate that for the majority of the cases, our proposed RccGAN outperforms other conventional and modern generative models in terms of accuracy, stability, and reliability \footnote{The supplementary materials including speech signals are available at:\\ https://github.com/EsmaeilpourMohammad/RccGAN.git.}.   
\end{abstract}

%%%%%%%%%%%%%%%%%%%%%%%%%%%%%%%%%%%%%%%%%%%%%%%%%%%%%%%%%%%%%%%%%%%%

%%%%%%%%%%%%%%%%%%%%%%%%%%%%%%%%%%%%%%%%%%%%%%%%%%%%%%%%%%%%%%%%%%%%

% \fi
\IEEEpeerreviewmaketitle

%%%%%%%%%%%%%%%%%%%%%%%%%%%%%%%
\section{Introduction}
\label{sec:introduction}
%%%%%%%%%%%%%%%%%%%%%%%%%%%%%%%
\IEEEPARstart{O}{ver} the past years, the proliferation and huge success of the generative adversarial network (GAN) \cite{goodfellow2014generative} in the multimedia domains (e.g., image, audio, text, etc.), has highly encouraged researchers to benchmark such generative models on another sophisticated realm, namely tabular databases (i.e., each database might contain multiple tables) \cite{bourou2021review}. These tables often contain a large combination of continuous, discrete, and binary (boolean) records with fairly low correlation (i.e., they are heterogeneous \cite{sheth1990federated,wang1990polygen}) compared to homogeneous databases in the multimedia domains \cite{borisov2021deep}.  According to the literature, the recent devoted attention to developing GANs for tabular data is also related to the demands of many companies which maintain such massive databases, for instance financial institutions \cite{even2007economics}, networking companies \cite{kayumovich2019capability}, healthcare organization \cite{trifiro2018big}, etc. It has been demonstrated that GANs can efficiently contribute to the benefit of these corporations and facilitate the process of mining critical information from large-scale tabular databases in different aspects. \cite{xu2020synthesizing}. More specifically, GANs can be used for augmenting sparse tables of the transactional databases for improving the aforementioned mining procedure \cite{engelmann2020conditional}, contribute to exploiting semantic relations among tables for marketing purposes \cite{yang2015semantic}, and protecting private and confidential information of the clients \cite{xu2019modeling}. Our focus in this paper is on the latter aspect since data leakage has always been a mounting concern for companies which collect personal information about their customers \cite{papadimitriou2010data}. 

Explaining the application of GANs for securing information of tabular databases is fairly straightforward. Assuming the personal information of clients in an organization are maintained in table $\mathcal{T}_{org}$ and it should be internally shared among employees in different departments such as marketing, finance, customer service, research and development, etc. For avoiding any potential data leakage, it is possible to train a GAN and synthesize a new table as $\mathcal{T}_{syn}$ according to the following reliability conditions: 
\begin{enumerate}[(a)]
\item The statistical properties including probability distribution of $\mathcal{T}_{syn}$ and $\mathcal{T}_{org}$ should be very similar,
\item The synthesized records (rows in a table) should be unanimously non-identical to the associated original records,
\item No original record should be identifiable from $\mathcal{T}_{syn}$.
\end{enumerate}

\noindent Finally, sharing $\mathcal{T}_{syn}$ instead of $\mathcal{T}_{org}$ with other departments not only is safe, but also facilitates the feasibility of conducting joint collaborations with external partners.

On the other hand, developing a stable GAN which can meet the aforementioned conditions is very challenging mainly because tabular databases are inherently heterogeneous. This obliges to employ efficient data preprocessing techniques in order to transform records into a more correlated representation. In fact, the convergence of the entire GAN is highly dependent to this representation. Moreover, the architecture of the GAN should be carefully designed to make a reasonable balance between accuracy and optimality in terms of the total number of required training parameters. Towards reaching to such a balance for the generative model we make the following contributions in this paper:
\begin{enumerate}[(i)]
\item Developing a new configuration for the tabular GAN with one generator and the possibility of employing multiple discriminator networks; 
\item Introducing the compound conditional generative model to better control over the consistency of the synthesized records; 
\item Proposing a novel regularization technique for training the GAN which is basically designed to address the model's instability issue (i.e., exploding the weight vectors or the learning curve of the loss function at certain iterations) during training; 
\item Characterizing a new metric for monitoring the stability of the generator network in order to precisely identify onset of instabilities;
\item Evaluating the performance of the GAN on six public benchmarking datasets using three statistical metrics to measure the aforementioned conditions. 
\end{enumerate}
The rest of the paper is organized as the following. Section~\ref{sec:background} briefly reviews the state-of-the-art GAN configurations which have been developed for synthesizing large-scale tabular datasets. Section~\ref{sec:proposedMethod} provides details of our proposed generative model as well as the explanation on the compound conditional GAN and the regularization scheme. Section~\ref{sec:experiments} summarizes our achieved results on some tabular databases from different categories. Finlay, Section~\ref{sec:discussions} presents some discussions about peripheral aspects of our proposed tabular data synthesis approach.

%%%%%%%%%%%%%%%%%%%%%%%%%%%%%%%%%%%%%%%%%%%%%%%%%%%%%%%%%
\section{background}
\label{sec:background}
A typical configuration of a GAN employs two deep neural networks known as the generator and discriminator represented by $\mathcal{G}(\cdot)$ and $\mathcal{D}(\cdot)$, respectively \cite{goodfellow2014generative}. The input of $\mathcal{G}(\cdot)$ is the vector $\mathbf{z}$ that is usually drawn from a probability distribution $p_{z} \sim \mathcal{N}(0,\mathbf{I})$ and this network is supposed to output (synthesize) samples very similar to the originals. This similarity is measured through comparing the probability distribution of such synthetic outputs and the real samples denoted by $p_{g}$ and $p_{r}$, respectively. However, the control of this measurement procedure is on the discriminator which provides gradients for updating the weight vectors of $\mathcal{G}(\cdot)$. Towards this end, many loss functions have been introduced for $\mathcal{D}(\cdot)$ which mostly rely on the basic cross-entropy formulation \cite{goodfellow2014generative}:
\begin{equation}
    \mathcal{L}_{\mathcal{D}} \equiv  \mathbb{E}_{\mathbf{z} \sim \mathcal{N}(0,\mathbf{I})}\log\left [ 1-\mathcal{D}(\mathcal{G}(\mathbf{z})) \right ] - \mathbb{E}_{\mathbf{x} \sim \mathbf{X}}\log \mathcal{D}(\mathbf{x})
    \label{eq:loossDisc}
\end{equation}
\noindent where $\mathbf{z}_{i} \in \mathbb{R}^{d_{z}}$ with dimension $d_{z}$ is a random vector and $\mathbf{x}$ stands for an original sample drawn form the real dataset, $\mathbf{X}$. Since $\mathcal{L}_{\mathcal{D}}$ is developed for binary classification problems, the output logit of $\mathcal{D}(\cdot)$ is either 0 or 1. Unfortunately, these discrete values might negatively affect the performance of the generator through imposing instability and mode collapse (i.e., losing sample variations and memorizing a limited number of modes) issues \cite{arjovsky2017wasserstein,xu2019modeling}. For tackling these potential issues and improving the generalizability and accuracy of the GAN, Arjovsky {\it et al.}~\cite{arjovsky2017wasserstein} proposed the critic function trick which is another variant for Eq.~\ref{eq:loossDisc} as the following.
\begin{equation}
    \hat{\mathcal{L}}_{\mathcal{D}} \equiv  \mathbb{E}_{\mathbf{z} \sim \mathcal{N}(0,\mathbf{I})}\log\left [ f_{c}(\mathcal{G}(\mathbf{z})) \right ] - \mathbb{E}_{\mathbf{x} \sim \mathbf{X}}\log f_{c}(\mathbf{x})
    \label{eq:lossDD}
\end{equation}
\noindent where $f_{c}(\cdot)$ represents the critic function and enables yielding continuous logits for $\hat{\mathcal{L}}_{\mathcal{D}}$. More specifically, $f_{c}(\cdot)$ measures the statistical discrepancy between samples drawn from $p_{g}$ and $p_{r}$ via optimizing for the following integral probablity metric (IPM) \cite{muller1997integral,sriperumbudur2012empirical}:
\begin{equation}
    \sup_{f_{c}\in \mathcal{F}} \left [ \mathbb{E}_{\mathcal{G}(\mathbf{z}_{i}) \sim p_{g}} f_{c}(\mathcal{G}(\mathbf{z}_{i}))- \mathbb{E}_{\mathbf{x} \sim p_{r}} f_{c}(\mathbf{x}) \right ]
    \label{eq:discrepany}
\end{equation}
\noindent where, $\mathcal{F}$ indicates the function class (e.g., the conjugacy classes) \cite{serre1977linear,sriperumbudur2012empirical,frasin2011new} and it is independent to the neural networks, $p_{g}$, and $p_{r}$. For effectively controlling over the critic function, often some restrictions are applied such as Lipschitz continuity in Wasserstein-GAN \cite{arjovsky2017towards} and Hilbert unit ball constraint in MMD-GAN \cite{li2017mmd} where MMD refers to the kernel maximum mean discrepancy. Toward achieving an accurate and comprehensive $f_{c}(\cdot)$, the generator network should be efficiently trained following a simple yet effective loss function, namely \cite{arjovsky2017towards}:
\begin{equation}
    \mathcal{L}_{\mathcal{G}} \equiv - \mathbb{E}_{\mathbf{z} \sim \mathcal{N}(0,\mathbf{I})}f_{c}(\mathcal{G}(\mathbf{z})), \quad  f_{c}(\mathcal{G}(\mathbf{z})) \approx \log \mathcal{D}(\mathcal{G}(\mathbf{z}))
    \label{eq:losssG}
\end{equation}
\noindent where $0 \leq \mathcal{D}(\mathcal{G}(\mathbf{z})) \leq 1$. Recently, substantial improvements have been constituted for developing the loss functions of both the generator and discriminator networks\cite{wang2018improving,mroueh2017fisher}. Moreover, various regularization techniques \cite{kurach2018gan} in addition to complex configurations have been introduced. These advanced configurations include residual \cite{gnanha2022residual}, attention-based \cite{chen2018attention}, and class-conditioned \cite{mirza2014conditional} designs. In the following, we review such a state-of-the-art GANs which have been mainly developed for tabular databases with a major focus on the synthesis platform.

$\mathrm{MedGAN}$ \cite{choi2017generating} is amongst the baseline generative models which have been developed for synthesizing large-scale tabular databases. The framework of this approach is synthesizing medical tables which include complex combinations of multimodal records. The great novelty of $\mathrm{MedGAN}$ is employing an autoencoder on top of the generator in order to improve the training accuracy of the model. This autoencoder exploits a different and independent architecture compared to both $\mathcal{G}(\cdot)$ and $\mathcal{D}(\cdot)$ networks. Furthermore, it implements a more straightforward optimization formulation as follows.     
\begin{equation}
    \min \sum \mathbf{x} \log \hat{\mathbf{x}}+(1-\mathbf{x})\log (1-\hat{\mathbf{x}}), \quad \hat{\mathbf{x}}=\mathrm{dec}(\mathrm{enc}(\mathbf{x})) 
\end{equation}
\noindent where $\mathrm{enc}(\cdot)$ and $\mathrm{dec}(\cdot)$ refer to the encoding and decoding functions of the autoencoder, respectively. During the training phase, $\mathrm{MedGAN}$ transforms the input records into a lower dimensional vector space using $\mathrm{enc}(\cdot)$ and then forces the generator to learn from $\hat{\mathbf{x}}_{i}$s. Thus, to comply with this setup, the discriminator is assigned to measure the similarity between $\mathrm{dec}(\mathcal{G}(\mathbf{z}_{i}))$ and $\mathbf{x}_{i} \in \mathbf{X}$. From the computational complexity point of view, this configuration is significantly costly, particularly in term of the total number of required training parameters. However, the autoencoding procedure provides a wider learning space for the generator by returning more informative gradients from $\mathcal{D}(\cdot)$ in every batch. 

$\mathrm{MedGAN}$ has been experimented on three public tabular datasets and it has shown a great performance in synthesizing binary and continuous records. For measuring the quality and accountability of the synthesized tables, some front-end classifiers have been employed. Cross-examining the recognition accuracy of such classifiers separately trained on $\mathcal{T}_{syn}$ and $\mathcal{T}_{org}$ corroborates the remarkable performance of $\mathrm{MedGAN}$ for large-scale tabular databases.

The second cutting-edge generative model which we review in this section is called $\mathrm{TableGAN}$ \cite{park2018data} which is computationally more efficient than the aforementioned $\mathrm{MedGAN}$. Moreover, it has relatively higher performance in runtime and it completely supports discrete records. Technically, $\mathrm{TableGAN}$ is an adapted version of the popular deep convolutional GAN (DCGAN) \cite{radford2015unsupervised} which has been extensively used in the multimedia domain, particularly for image synthesis and translation \cite{isola2017image,zhu2017unpaired}. Since records of the tabular datasets are 1D vectors, they should be accurately reshaped into squared matrices to fit the first layer of $\mathcal{G}(\cdot)$ according to the DCGAN configuration. This reshaping procedure obliges the generator to incorporate an additional network which is called the classifier $\mathcal{C}(\cdot)$ for more effectively controlling over the outputs of the generator and remove the potential outlier synthetic records. The architecture of such a network is identical to the discriminator with an adjusted loss function as the following.
\begin{equation}
    \min \mathbb{E}  \Big[ \left | \mathcal{V} \left ( \mathbf{x} \right )- \mathcal{C} \left ( \mathcal{R} \left ( \mathbf{x} \right ) \right ) \right |  \Big] \quad \mathrm{and} \quad  \forall \mathbf{x} \in \mathbf{X}
    \label{eq:tableGAN}
\end{equation}
\noindent where $\mathcal{V}(\cdot)$ retrieves and validates the accuracy of the real record in every batch and $\mathcal{R}(\cdot)$ denotes the removing function from $\mathcal{T}_{syn}$. This formulation is designed to detect potential synthetic records which do not match the distribution of the real ground-truth samples. 

$\mathrm{TableGAN}$ has been benchmarked on four public tabular repositories and the conducted investigations over the marginal distribution of $p_{g}$ relative to $p_{r}$ demonstrates the power of this generative model for synthesizing challenging databases.

Another generative model which has been extensively experimented for large-scale tabular data synthesis is $\mathrm{MotGAN}$ developed by Mottini {\it et al.} \cite{mottini2018airline}. The major difference between this generative model and the two aforementioned GANs is two-fold. Firstly, the generator network in $\mathrm{MotGAN}$ is based on optimizing for the squared energy distance \cite{rizzo2016energy} which is also known as the Cram\'{e}r IPM \cite{bellemare2017cramer}. The function class ($\mathcal{F}$) in this integral probability metric is complex and imposes a strict constraints on the generator as \cite{mroueh2017sobolev}:
\begin{equation}
    \mathbb{E}_{\mathbf{x}\sim p_{r}}\left [ \frac{\partial f_{c}(\cdot)}{\partial \mathbf{x}} \right ]^{2}\leq 1
\end{equation}
\noindent where $f_{c}$ is smooth with zero boundary condition. This constraint contributes to the stability of the generator and yields a more comprehensive model.

Secondly, the discriminator configuration in $\mathrm{MotGAN}$ enquires costly preprocessing operations such as token embedding particularly for discrete records. Moreover, the designed loss function for $\mathcal{D}(\cdot)$ is regularized using the gradient penalty technique ($\mathrm{GP}$) \cite{gulrajani2017improved} as follows.
\begin{equation}
    \mathcal{L}_{\mathcal{D}}= -\mathcal{L}_{\mathcal{G}}+\lambda\mathrm{GP}(\cdot)
\end{equation}
\noindent where $\lambda$ is a static nonzero scalar. For evaluating the performance of such a generative model on the tabular databases, the marginal similarity between $p_{g}$ and $p_{r}$ has been measured using the conventional Jensen-Shannon divergence (JSD) metric. The reported results demonstrate considerably high similarity between these two probability distributions in runtime. However, there is no reported discussion on the stability and the potential mode collapse issues during training. Addressing these concerns is deeply critical since unstable GANs with underrepresented modes might still result in high similarity between synthetic and real probability distributions \cite{srivastava2017veegan}. 

$\mathrm{VeeGAN}$ which stands for the variational encoder enhancement GAN \cite{srivastava2017veegan} has been primarily developed for tackling instability and mode collapse issues during training. This generative model employs a reconstructor network $\mathrm{Rec}(\cdot)$ on top of the generator to recover underrepresented modes. On one hand, $\mathrm{VeeGAN}$ is similar to $\mathrm{MedGAN}$ since they both incorporate an autoencoding policy for reconstructing samples. On the other hand, they are fundamentally different since unlike $\mathrm{MedGAN}$, the cross-entropy loss in $\mathrm{VeeGAN}$  involves the distributions of $\mathbf{z}_{i}$s as the following:

\begin{equation}
    \min \mathbb{E}\left [ \left \| \mathbf{z}-\mathrm{Rec}\left ( \mathcal{G}(\mathbf{z}) \right ) \right \|_{2}^{2} \right ]+\mathcal{O}\left ( \mathbf{z},\mathrm{Rec}(\mathbf{x}) \right )
\end{equation}
\noindent where $\mathcal{O}(\cdot)$ is another cross-entropy loss for measuring the relative discrepancy between the real and synthetic records. The architecture of the reconstructor is not necessarily dependent to neither the generator nor the discriminator network. However, $\mathrm{Rec}(\cdot)$ shares a similar configuration and settings with $\mathcal{D}(\cdot)$ to avoid imposing unnecessary computational overhead to the entire GAN. $\mathrm{VeeGAN}$ has been evaluated on a few public databases and it has demonstrated a remarkable performance both in terms of synthesis accuracy and improving model stability during training. Nevertheless, it only works with continuous records and it is not easily generalizable to discrete and binary samples.

One similar approach to $\mathrm{VeeGAN}$ in terms of employing the reconstructor network based on the autoencoding policies is called $\mathrm{InvGAN}$ which stands for the invertible tabular GAN \cite{lee2021invertible}. This generative model employs two stacked generator networks bundled inversely in such a way that the first $\mathcal{G}(\cdot)$ receives the latent variables of the autoencoder and reconstructs $\mathbf{z}_{i}$. Then, the second generator maps this derived vector into the synthetic record. This procedure filters the strongly correlated latent variables and avoids confusing $\mathcal{D}(\cdot)$ during training. According to the carried out experiments on several public databases, $\mathrm{InvGAN}$ is highly capable to synthesize large-scale tables with various modalities. 

The sixth generative adversarial model which explored in this section is the incomplete table synthesis GAN ($\mathrm{ItsGAN}$) \cite{chen2019faketables}. This advanced model has been primarily developed for data augmentation purposes applicable on massive databases containing sparse tables and records. In order to provide more informative gradients to the generator network during training, $\mathrm{ItsGAN}$ employs multiple autoencoding blocks. These blocks are pretrained and they share an identical architecture. In terms of functionality, the autoencoding blocks are significantly different from the autoencoder implemented for $\mathrm{VeeGAN}$. More specifically, the autoencoders in $\mathrm{ItsGAN}$ are sequentially organized into the discriminator module without any direct connection to the generator. This stack of autoencoders considerably improves the accuracy of the entire GAN and contributes to the stability of the model during training. The loss function of such autoencoders is the conventional cross-entropy with no regularization scheme.

$\mathrm{ItsGAN}$ forces the generator to learn functional dependencies (FD) from $\mathcal{T}_{org}$. FDs are analytic functions similar to the critic function in Eq.~\ref{eq:discrepany}. However, they do not belong to any function class $\mathcal{F}$. Toward this end, $\mathcal{G}(\cdot)$ is assigned to implicitly extract FDs during training. Since similar continuous columns in $\mathcal{T}_{org}$ most likely constitute indistinguishable FDs, providing accurate gradients to the generator network might be extremely challenging for the discriminator and increases its computational complexity. Regarding this concern, making the generator conditional to single or multiple columns might be considered a conservative approach and it is the basis idea behind introducing the class-conditional tabular GAN ($\mathrm{CctGAN}$) \cite{xu2019modeling}.

$\mathrm{CctGAN}$ is based on the standard conditional GAN which has been developed by Mirza {\it et al.}~\cite{mirza2014conditional} for image synthesis purposes. This category of GANs encodes labels of the classes into the input vector of the generator network providing additional background information about the training samples. This operation helps to avoid skipping underrepresented modes and improves stability of both the generator and discriminator networks. Since records of the tabular databases do not usually have an accompanying label, designing an accurate conditional vector is extremely critical. The definition of this vector for $\mathrm{CctGAN}$ is as the following.
\begin{equation}
\overrightarrow{\mathrm{cnd}}_{\mathrm{CctGAN}}=\mathbf{m}_{1}\oplus \mathbf{m}_{2}\cdots \oplus 
 \mathbf{m}_{n_{d}},  \quad  \mathbf{m}_{i}=\left [ \mathbf{m}_{i}^{(k)} \right ]_{k=1}^{n_{d}}
 \label{eq:conVecvec}
\end{equation}
\noindent where $k$ and $n_{d}$ denote the column index and the total number of discrete columns in the training table $\mathcal{T}_{org}$, respectively. Additionally, $\mathbf{m}_{i}$s are the mask functions containing a sequence of binary values. Every $\mathbf{m}_{i}^{(k)}$ holds the value of $1$ once the $i$-th column in $\mathcal{T}_{org}$ is randomly selected to be included in the input vector of $\mathcal{G}(\cdot)$, otherwise it holds zero. It has been demonstrated that making the generator conditional to $\overrightarrow{\mathrm{cnd}}_{\mathrm{CctGAN}}$ significantly improves the generalizability and stability of the model in runtime. 

$\mathrm{CctGAN}$ is quite compatible with any single or multimodal combination of continuous, binary, and discrete columns with various lengths. However, the values under every column of $\mathcal{T}_{org}$ should be transformed into consistent vector representations before the training. Implementing such a transformation is extremely important since it is designed to improve the performance of the model both in terms of accuracy and stability. For discrete and binary columns, $\mathrm{CctGAN}$ exploits the straightforward $\tanh(\cdot)$ transformer function which maps every value into the range $\left [ -1,1 \right ]$. However, for continuous columns it employs the variational Gaussian mixture model (VGM) \cite{bishop2007pattern} approach with different $m_{c}$ modes as the following.
\begin{equation}
    \sum_{k_{c}=1}^{m_{c}}\mu_{k_{c}}\mathcal{N}(c_{i,j}; \eta_{k_{c}},\sigma_{k_{c}})
    \label{eq:preproceCTC}
\end{equation}
\noindent where $c_{i,j}$ denotes a field under the column $C_{i}$. Every mode in the VGM is represented by a scalar value $\eta$ and a normal distribution with two parameters $\mu$ and $\sigma$ which indicate the mean and standard deviation, respectively. Since samples with short skewness might be discarded in Eq.~\ref{eq:preproceCTC} \cite{friedman2017elements}, an improved version of such formulation has been introduced in \cite{esmaeilpour2021bi}:
\begin{equation}
\sum_{k_{c}=1}^{m_{c}}\hat{\mu}_{k_{c}} \bigg[ \mathcal{N}\Big(c_{i,j}\mid \mu_{k_{c}},\sigma_{k_{c}}\Big) \bigg], \quad \hat{\mu}_{k_{c}} \sim  \mathcal{N}(0,0.5I)
\label{eq:bGAN}
\end{equation}
\noindent where $\hat{\mu}_{k_{c}}$ redistributes every column $C_{i}$ around $\mathcal{N}(\mu_{k_{c}}, \sigma_{k_{c}})$ and enables the VGM to yield a more accurate representation for the continuous values.

The efficiency of such a transformation technique has been fully investigated in a similar conditional generative model platform and it is called the bi-discriminator conditional tabular GAN ($\mathrm{BctGAN}$) \cite{esmaeilpour2021bi}. This synthesis approach exploits two discriminator networks for a single and fully convolutional generator network. Averaged over four public large-scale tabular databases and at the cost of tripling the number of training parameters, $\mathrm{BctGAN}$ outperforms the aforementioned synthesis approaches. Additionally, it dominantly surpasses the conventional generative models such as the standard classical Bayesian network ($\mathrm{CLBN}$) \cite{chow1968approximating} and the well known private Bayesian network ($\mathrm{PrivBN}$) \cite{zhang2017privbayes}. However, $\mathrm{BctGAN}$ also suffers from instability and mode collapse issues at larger iterations which forces to early stop the training procedure.

In the following section, we explain our proposed approach for tabular data synthesis which addresses all the abovementioned raised concerns about generative models such as accuracy, generalizability, stability, and mode collapse.

%%%%%%%%%%%%%%%%%%%%%%%%%%%%%%%%%%%%%%%%%%%%%%%%%%%%%%%%%

%%%%%%%%%%%%%%%%%%%%%%%%%%%%%%%%%%%%%%%%%%%%%%%%%%%%%%%%%
\section{Proposed Method: Regularized Compound Conditional GAN ($\mathrm{RccGAN}$)}
\label{sec:proposedMethod}
We develop a conditional GAN for synthesizing large-scale tabular databases. Our motivation behind employing the class-conditional platform over the conventional and regular configurations is enhancing the accuracy and improving the stability of the generative model during training. Since learning multimodal values with low correlation is usually challenging for GANs \cite{zhang2019multi}, conditioning the generator to a single or multiple vectors helps yielding a more robust model \cite{esmaeilpour2021bi}. There are four major concepts relevant to designing our synthesis approach which we will separately discuss them in the following subsections.

%%%%%%%%%%%%%%%%%%%%%%%%%%%%%%%%%%%%%%%%%%%%%%%%%%%%%%%%%
\subsection{Tabular Data Preprocessing}
\label{sec:subsecpreproc}
Every record in $\mathcal{T}_{org}$ accommodates a number of fields in which every field is associated with a column. Not only such columns can be different in terms of modality (i.e., format), but also they might have various length and scales. Therefore, tables might be extremely heterogeneous and this discrepancy negatively affects the performance of the learning algorithms during training. This is fairly intuitive since weakly correlated records limitate the learning space of any data-driven model and consequently result in lower recognition accuracy. Presumably, the straightforward approach to address this issue is transforming either records or columns into more correlated vector spaces and normalize their distributions. On the other hand, such normalization operations can be also challenging since they might increase the chance of memorization for the learning algorithms \cite{esmaeilpour2021bi}.   

Following $\mathrm{CctGAN}$, we also use $\tanh(\cdot)$ mapping function for encoding discrete and binary columns into normalized one-hot vectors \cite{li2018clustering}. However, we design a novel VGM-based transformer for continuous columns to better represent values with multivariate distributions. Toward this end, we train a VGM on every continuous column and fit $m_{c}$ Gaussian models (in a mixture setup \cite{huang2017model}) with parameters $\left \langle \mu_{k_{c}}, \sigma_{k_{c}} \right \rangle$. Assuming the combination of these models form the mixing probability distribution $\varphi (\cdot)$, then using the normal variance-mean mixture (NVMM) theorem \cite{barndorff1982normal} we can redistribute the paired parameters $\left \langle \mu_{k_{c}}, \sigma_{k_{c}} \right \rangle$ over the universal distribution $\varphi (\cdot)$ as follows.
\begin{equation}
\int_{0}^{\infty}\frac{1}{\sqrt{2\pi\sigma^{2}_{k_{c}}v}}\exp{\left ( \frac{-(\mu_{k_{c}}-\alpha-\beta v)^{2}}{2\sigma^{2}_{k_{c}} v} \right )\varphi (v)dv} 
\label{eq:nvmmt}
\end{equation}
\noindent where $v$ is a random variable. Additionally, $\alpha$ and $\beta$ denote real values associated with the combination of $m_{c}$ Gaussian modes. This redistribution operation translates the paired parameters $\left \langle \mu_{k_{c}}, \sigma_{k_{c}} \right \rangle$ into \cite{luciano2010generalized}:
\begin{equation}
    \left \langle \underbrace{\alpha+\beta \omega_{k_{c}}}_{\tilde{\mu}_{k_{c}}} , \underbrace{\sigma_{k_{c}} \omega_{k_{c}}}_{\tilde{\sigma}_{k_{c}}} \right \rangle, \quad \omega_{k_{c}} \sim \varphi
    \label{eq:redistriparam}
\end{equation}
\noindent where $\tilde{\mu}_{k_{c}}$ and $\tilde{\sigma}_{k_{c}}$ stand for the mean and standard deviation parameters of the translated VGM modes, respectively. The intuition behind this reparameterization is two fold. Firstly, suppressing the Gaussian models with high variances which dominate other low-profile modes. Secondly, having a full control over the VGM modes through manipulating $\alpha$ and $\beta$. This contributes to effectively tune the correlation among the $m_{c}$ modes. In order to find optimal values for both $\alpha$ and $\beta$, we do not employ any moment generating function to avoid adding unnecessary complications to the preprocessing operation. In fact, we assume these variables are randomly drawn from the uniform distribution $\mathcal{U}\left [ -1,1 \right ]$. However, we impose a uniformity constraint on Eq.~\ref{eq:nvmmt} to avoid memorizing $\mu_{k_{c}}$s derived from the original VGM model as follows.
\begin{equation}
\left | \alpha+\beta \omega_{k_{c}} - \mu_{k_{c}} \right |_{2} \geq \epsilon
\label{eq:alphabetaepsil}
\end{equation}
\noindent where $\epsilon$ is a non-zero scalar and it should be defined according to the properties of the achieved VGM modes such as marginal likelihood \cite{constantinopoulos2006bayesian}. This formulation enables us to represent every field of each column relative to $\omega_{k_{c}}$ as:
\begin{equation}
    p_{c_{i,j}} \equiv \omega_{k_{c}}\mathcal{N} \Big ( c_{i,j} \mid \tilde{\mu}_{k_{c}}, 
    \tilde{\sigma}_{k_{c}} \Big )
    \label{eq:pciandj}
\end{equation}
\noindent where $p_{c_{i,j}}$ describes the probability of the field $c_{i,j}$ coming from the reparameterized modes. If $\epsilon$ is too large, then $p_{c_{i,j}}$ tends to zero and result in an ill-conditioned density function \cite{moghavvemi2001technique}. Therefore, fine-tuning $\epsilon$ is a critical step toward accurately representing continuous values.      

All the above-mentioned data preprocessing operations have been designed to be compatible and useful for implementing any configurations of conditional GANs. More specifically, these operations help to simplify the definition of conditional vectors. Additionally, this simplification contributes to improving the computational complexity and accuracy of the entire generative model compared to the $\mathrm{CctGAN}$ and its variants (e.g., $\mathrm{BctGAN}$). We provide a complete discussion about defining such conditional vectors in the following subsection.

%%%%%%%%%%%%%%%%%%%%%%%%%%%%%%%%%%%%%%%%%%%%%%%%%%%%%%%%%

%%%%%%%%%%%%%%%%%%%%%%%%%%%%%%%%%%%%%%%%%%%%%%%%%%%%%%%%%
\subsection{Designing the Compound Conditional Vector}
\label{sec:subCondVec}
Accurately defining the conditional vector helps to prepare a comprehensive learning space for the generator network during training \cite{brock2018large,li2019storygan}. Therefore, such a definition becomes more critical for tabular GANs regarding the heterogeneous nature of $\mathcal{T}_{org}$. One of the common approaches for deriving a valid conditional vector is using the mask functions similar to Eq.~\ref{eq:conVecvec} \cite{xu2019modeling}. There are two major side effects for this type of formulation. Firstly, mask functions only incorporate the index of discrete columns in crafting $\mathbf{m}_{i}$s and do not involve binary features mainly for the sake of simple implementation \cite{xu2019modeling}. Secondly, the dimension of $\mathbf{m}_{i}$s progressively increases with the cardinality of discrete columns. In other words, for large-scale tables with numerous discrete columns, $\mathbf{m}_{i}$s becomes a long-tailed sequence. Thus, training the generator conditioned to such a lengthy vector might be extremely challenging. In address to these two concerns, we introduce a new formulation for the conditional vector without using any mask function.

For involving actual values of discrete and binary columns into the definition of the conditional vectors, we fit a Cantor distribution $\mathcal{CD}$ \cite{graf1997quantization} separately on their associated one-hot vectors. Our intention for employing $\mathcal{CD}$ originates from the unique statistical characteristics of this distribution which is leveraging between probability density and mass functions \cite{ben1981non}. More specifically, $\mathcal{CD}$ is neither absolutely discrete nor continuous and consequently it complies with the nature of both binary and discrete columns. Moreover, since there is no correlation between $\varphi$ and $\mathcal{CD}$ in terms of statistical properties, the generator network can alleviate the chance of memorizing sample distributions. In fact, the discrepancy between these two distributions helps to avoid deriving biased conditional vectors.

The straightforward approach for obtaining parameters of the Cantor distribution is approximating its moment generating function as:
\begin{equation}
    \mathcal{CD}_{t} \approx  e^{t/2}\prod_{k_{t}=1}^{\infty}\cosh \left ( t\cdot 3^{-k_{t}} \right )
    \label{eq:cantormomentgen}
\end{equation}
\noindent where $t$ is a random variable which can be substituted with the variable of either discrete or binary columns represented by $d$ and $b$, respectively. Hence, after such a substitution, $k_{d}$ and $k_{b}$ individually denote the dimension of discrete and binary one-hot vectors. Finally, we can define our conditional vector as the following. 
\begin{equation}
    \overrightarrow{\mathrm{cond}} := \left [ \eta_{k_{d},d} \sim \mathcal{CD}_{d} \right ] \oplus \left [ \eta_{k_{b},b} \sim \mathcal{CD}_{b} \right ]
    \label{eq:condiGeni}
\end{equation}
\noindent where $\oplus$ refers to the mathematical concatenation operator. Additionally, $\eta_{d}$ and $\eta_{b}$ denote the discrete and binary vectors randomly drawn from $\mathcal{CD}_{d}$ and $\mathcal{CD}_{b}$, respectively. In the following subsection, we explain how to make the generator network conditional to Eq.~\ref{eq:condiGeni} and we provide an extensive discussion about the configuration of both $\mathcal{G}(\cdot)$ and $\mathcal{D}(\cdot)$.

%%%%%%%%%%%%%%%%%%%%%%%%%%%%%%%%%%%%%%%%%%%%%%%%%%%%%%%%%

%%%%%%%%%%%%%%%%%%%%%%%%%%%%%%%%%%%%%%%%%%%%%%%%%%%%%%%%%
\begin{figure*}[t]
  \centering
  \includegraphics[width=\textwidth]{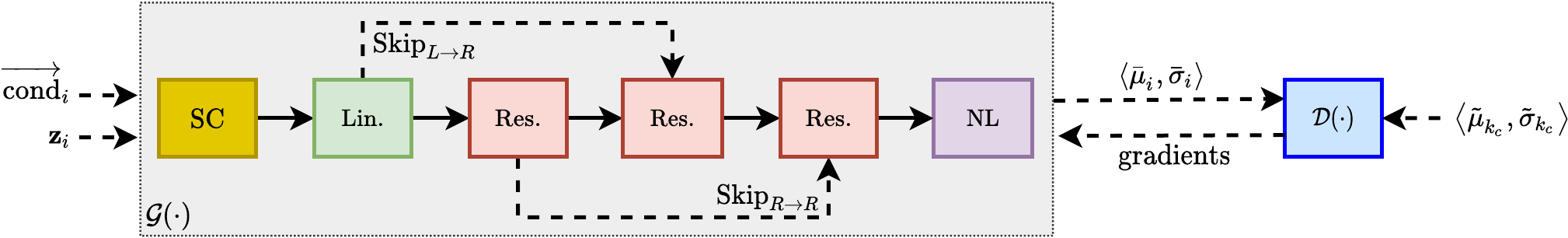}
  \caption{Overview of the proposed configuration for $\mathrm{RccGAN}$. The split and concatenation operations have been represented by $\mathrm{SC}$. Additionally, the linear, residual, and non-local blocks are indicated by $\mathrm{Lin.}$, $\mathrm{Res.}$, and $\mathrm{NL}$, respectively. Moreover, $\mathrm{Skip}_{\mapsto}$ refers to the skip connection from a source to another target block. The gray box highlights insides of the generator configuration.} 
  \label{overview-GANarchitt}
  %\vspace{-15pt}
\end{figure*}
\subsection{Configuration of the Proposed GAN}
\label{sec:subarchGAN}
The overview of our proposed configuration for $\mathrm{RccGAN}$ is depicted in Fig.~\ref{overview-GANarchitt}. As shown, the first processing module of the generator is split and concatenation ($\mathrm{SC}$) which is inspired from \cite{brock2018large}. Since there is no restriction on the dimensionality of $\mathbf{z}_{i}$s, $\mathrm{SC}$ divides every input vector into smaller fixed-length bins and concatenates them with the conditional vector. On one hand, this procedure contributes to avoid providing lengthy input vectors to the generator which might negatively affect the training performance of the entire GAN. On the other hand, there is always the risk of underrepresenting $\mathbf{z}_{i}$s by dividing them into smaller chunks. More specifically, this division operation usually enquires padding the achieved sub-vectors (bins of $\mathbf{z}_{i}$) with a bunch of zeros to fit the dimension of fixed-length bins. One potential approach to address this issue is empirically tuning $\mathrm{SC}$ through conducting exploratory experiments to achieve appropriate bins.  

The second processing module of the designed generator in Fig.~\ref{overview-GANarchitt} is the linear block \cite{brock2018large}. Technically, this block is a fully connected neural network with multiple hidden layers which projects the conditional and input vectors onto a sequence of parallel channels $\mathcal{CH}_{j}$ as:
\begin{equation}
    \mathrm{SC} \Big ( \overrightarrow{\mathrm{cond}}_{i}, \mathbf{z}_{i} \Big) \mapsto \mathcal{CH}_{j} \quad j \in \left \{ 1,2, \cdots, \varpi  \right \}
\end{equation}
\noindent where $\varpi$ is a hyperparameter and it should be empirically set during training. These channels uniformly share the learned parameters derived from the linear block without employing dropout policy. Half of such channels pass through the subsequent module which is the first residual block ($\mathrm{Res.}$) and the remaining $\mathcal{CH}_{j}$s form a skip connection to the second $\mathrm{Res.}$ represented by $\mathrm{Skip}_{L\rightarrow R}$ in Fig.~\ref{overview-GANarchitt}. There is a similar skip connection between the first and third residual block which is denoted by $\mathrm{Skip}_{R\rightarrow R}$. The motivation behind embedding $\mathrm{Skip}_{L\rightarrow R}$ into the generator configuration is restoring the distribution of the learned parameters from the linear module to the second residual block. This restoration policy is designed to counteract with the potential explosion of weight vectors mainly through the backpropagation phase from $\mathrm{Res.}$ to $\mathrm{Lin.}$ during training. Preventing such an explosion is crucial since it is closely related with the mode collapse and instability issues in GAN setups \cite{brock2017neural}.

Our intention for incorporating $\mathrm{Skip}_{R\rightarrow R}$ into the configuration of the generator is two fold. Firstly, avoiding the potential domination of $\mathrm{Skip}_{L\rightarrow R}$ on the performance of the residual blocks. Secondly, reinforcing the final residual block (the third $\mathrm{Res.}$) to resist against memorizing input vectors. Finally, we pass the achieved embeddings into the non-local block \cite{brock2018large,wang2018non} represented by $\mathrm{NL}$ in Fig.~\ref{overview-GANarchitt}. The output of this block is a Gaussian model with parameters $\bar{\mu}_{i}$ and $\bar{\sigma}_{i}$. In summary, we can formulate the functionality of the generator network as the following. 
\begin{equation}
\mathcal{G}\Big (\mathbf{z}_{i} \mid \overrightarrow{\mathrm{cond}}_{i} \Big ) = \left \langle {\bar{\mu}}_{i},\bar{\sigma}_{i} \right \rangle     
\end{equation}
\noindent where the precision of such model should be measured via the discriminator network. Although $\mathcal{D}(\cdot)$ in Fig.~\ref{overview-GANarchitt} is depicted as a singular network, it can be definitely generalized to multiple discriminators for returning more informative gradients to the generator.

Our proposed configuration for $\mathrm{RccGAN}$ is not dependent neither to the architecture of the generator nor the discriminator. In other words, each block in Fig.~\ref{overview-GANarchitt} can employ any number of hidden layers, various types of activation functions, different schemes for initialization, etc. Unfortunately, there is no analytical approach for finding an optimal architecture for both $\mathcal{G}(\cdot)$ and $\mathcal{D}(\cdot)$. In fact, designing such architectures are subjective and they may vary according to the properties of the benchmarking databases. Therefore, we introduce our preferred architectures in Section~\ref{sec:experiments}.   

%%%%%%%%%%%%%%%%%%%%%%%%%%%%%%%%%%%%%%%%%%%%%%%%%%%%%%%%%

%%%%%%%%%%%%%%%%%%%%%%%%%%%%%%%%%%%%%%%%%%%%%%%%%%%%%%%%%
\subsection{Regularization Protocol for the Generator}
\label{sec:subReguProtc}
This subsection explains details of a novel regularization scheme which we developed it for smoothly training the generator. Basically, this technique is inspired from the orthogonal regularization ($\mathrm{OR}$) approach \cite{brock2017neural} which uniformly normalizes the weight vectors of the generator network. This helps to partially bypass the sudden explosion of weight vectors during training. Since it has been proven that $\mathrm{OR}$ can be prohibitive in runtime \cite{miyato2018spectral}, we introduce a more straightforward technique using the Gershgorin Theorem \cite{van1983matrix}.

\noindent \textbf{Proposition:} Assuming $\theta_{g}$ represents the weight vector of the generator network. Then, it is possible to bound the variation of $\theta_{g}$ through its eigenvector within a nondeterministic interval as:
\begin{equation}
    \Big [ \mathrm{trace}(\theta_{g})-\vartheta,  \mathrm{trace}(\theta_{g})+\vartheta  \Big]
    \label{eq:intervalRegul}
\end{equation}
\noindent where $\mathrm{trace}(\cdot)$ is a basic algebraic function for computing sum of eigenvalues for the given vector \cite{van1983matrix} and $\vartheta$ is an error term. We prove the existence of such an interval as the following.  
\begin{figure*}[th]
  \centering
  \includegraphics[width=0.9\textwidth]{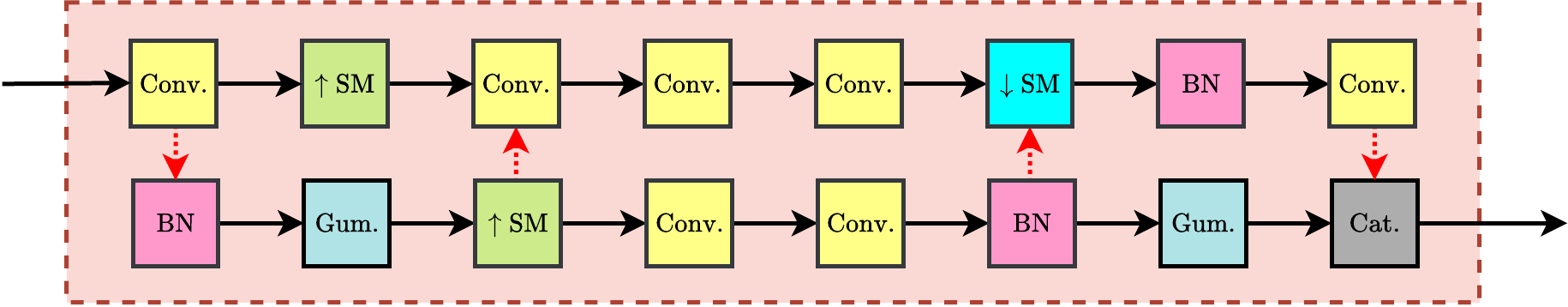}
  \caption{Overview of the designed network for the residual block in Fig.~\ref{overview-GANarchitt}. The convolution and batch normalization layers are shown by $\mathrm{Conv.}$ and $\mathrm{BN}$, respectively. Additionally, upsampling and downsampling operations with ratio of $0.5$ are indicated by $\uparrow \mathrm{SM}$ and $\downarrow \mathrm{SM}$. Furthermore, the Gumbel softmax and the concatenation layers are denoted by $\mathrm{Gum.}$ and $\mathrm{Cat.}$, sequentially. Herein, the iterative transitions from a block to another are depicted by red dotted arrows.} 
  \label{RccGAN-gen}
\end{figure*}

\noindent \textbf{Proof:} Supposing $\mathcal{Y}$ and $Q$ are random singular vectors defined in the Lebesgue vector space and $\lambda$ is the eigenvector associated with $\theta_{g}$. Thus, we can write \cite{van1983matrix}:
\begin{equation}
    1 \leq \underbrace{\left \| \Big ( \underbrace{ Q^\top\theta_{g}Q- \mathcal{Y}}_{\mathcal{H}}-\lambda I \Big)^{-1}\mathcal{Y} \right \|_{\infty}}_{\sum_{j_{o}=1}^{n_{o}} \left | y_{k,j_{0}} \right | \left |  \mathcal{H}_{k}-\lambda_{k} \right |^{-1}} \mathrm{and} \quad 1\leq k \leq n_{0}
    \label{eq:proofEqq}
\end{equation}
\noindent where $n_{0}$ refers to the maximum length of the eigenvector and $\mathcal{H}$ is a nonzero vector. Since $Q$ is orthogonal (based on the Gershgorin theorem \cite{stewart1975gershgorin}), thus significant variation in $\theta_{g}$ causes:
\begin{equation}
    \left | \mathcal{H}_{k}-\lambda_{k} \right |\rightarrow 0
\end{equation}
\noindent and consequently results in an ill-conditioned case for Eq.~\ref{eq:proofEqq}. Hence, $\lambda$ becomes analytical only within the following confidence interval \cite{van1983matrix}:
\begin{equation}
    \lambda_{k} \in \left [ \mathcal{H}_{k}-\lambda_{k} , \mathcal{H}_{k}+\lambda_{k}  \right ], \quad \forall k \leq n_{0}
\end{equation}
\noindent For large values of $k$, $\lambda_{k}\rightarrow 0$ and $\mathcal{H}_{k} \rightarrow \mathrm{trace}(\theta_{g})$ \cite{van1983matrix}. Statistically, there are some integrated analytical approaches for closely approximating $\vartheta$ such as implementing the least square error functions based on Wielandt-Hoffman theorem \cite{chu1991least}. However, such implementations add unnecessary computational overhead to the training procedure of the generator network. For tackling this issue, we empirically initialize $\vartheta$ and tune it according to the distribution of $\lambda_{k}$ as the following.
\begin{equation}
    \left | \vartheta  \right | \leq 1 + \mathrm{median}(\lambda)
    \label{eq:upperboundTra}
\end{equation}
\noindent where $\mathrm{median}(\cdot)$ indicates the conventional statistical function for computing the median of a given vector. We believe that such an upperbound is large enough to avoid dramatic changes on $\theta_{g}$.\hfill $\square$

From yet another perspective, imposing Eq.~\ref{eq:upperboundTra} on the generator network might result in late convergence for the entire GAN setup. We investigate this potential side effect in Section~\ref{sec:experiments}.

%%%%%%%%%%%%%%%%%%%%%%%%%%%%%%%%%%%%%%%%%%%%%%%%%%%%%%%%%

%%%%%%%%%%%%%%%%%%%%%%%%%%%%%%%%%%%%%%%%%%%%%%%%%%%%%%%%%
\section{Experiments}
\label{sec:experiments}
This section explains precise details of our conducted experiments and analyzes the achieved results on six public benchmarking databases. Additionally, it investigates the stability of $\mathrm{RccGAN}$ at large iterations during training and measures the compliance of the proposed model with the three conditions mentioned in Section~\ref{sec:introduction}.

\subsection{Databases}
\label{sec:databasesub}
We carry out our experiments on six tabular databases which have been previously benchmarked for synthesis purposes \cite{xu2019modeling}. More specifically, we selected three databases from the public UCI repository \cite{dua2017uci}, namely Adult, Census, and News which contain thousands of continuous, discrete, and binary records with various length and scales. Additionally, we selected \href{https://www.kaggle.com/mlg-ulb/creditcardfraud}{Home Credit Default Risk (HCDR)}, \href{https://www.kaggle.com/arashnic/taxi-pricing-with-mobility-analytics}{Cabs}, and \href{https://www.kaggle.com/harlfoxem/housesalesprediction}{King} databases from the Kaggle machine learning archive. The importance of employing these databases for evaluating the performance of the synthesis approaches has been characterized in \cite{lee2021invertible}. For instance, such databases include multivariate records associated with multimodal columns. Therefore, they can effectively challenge different tabular data synthesis algorithms and identify their potential pitfalls. Furthermore, Adult, Census, HCDR, and Cabs are essentially collected for classification and the rest for regression problems. Hence, we can compare the performance of the tabular GANs for two major machine learning tasks.

Since the aforementioned databases are not normalized, we firstly transform them into a more correlated vector space as explained in Section~\ref{sec:subsecpreproc}. For every continuous column, we train a separate VGM model with $m_{c}=10$ and empirically set $\epsilon$ to $0.005$ for Eq.~\ref{eq:alphabetaepsil}. On one hand, assigning constant values to these hyperparameters for six different databases might not be absolutely optimal. On the other hand, it is to the benefit of the computational complexity since we no longer need to heuristically search for both $m_{c}$ and $\epsilon$. Finally, if the overall performance of the mixture models in the VGMs is not satisfactorily high, we can fine-tune the associated parameters using the grid-search policy \cite{bardenet2013collaborative}.

In the following subsection, we design the architectures of both the generator and discriminator networks for training them on the normalized tables.

\begin{table*}[th]
\centering
\scriptsize
\caption{Comparison of averaged $F_{1}$ scores for the front-end classifiers trained on the synthesized tables using different generative models relative to the ground-truth (original database). We run 5-fold cross validation for all the tables and unanimously configure the ratio of the validation and test subsets to 0.7 and 0.3, respectively. Outperforming results are shown in bold-face.}
\resizebox{\textwidth}{!}{\begin{tabular}{r||cccccccc}
\hline
\multicolumn{1}{c||}{\multirow{3}{*}{Generative Model}} & \multicolumn{8}{c}{Tabular Database for the classification tasks}                                                                                                                                                                                                                                                                     \\ \cline{2-9} 
\multicolumn{1}{c||}{}                        & \multicolumn{2}{c||}{Adult}                                                    & \multicolumn{2}{c||}{Census}                                                   & \multicolumn{2}{c||}{HCDR}                                                     & \multicolumn{2}{c}{Cabs}                                 \\ \cline{2-9} 
\multicolumn{1}{c||}{}                        & \multicolumn{1}{c|}{Validation}       & \multicolumn{1}{c||}{Test}             & \multicolumn{1}{c|}{Validation}       & \multicolumn{1}{c||}{Test}             & \multicolumn{1}{c|}{Validation}       & \multicolumn{1}{c||}{Test}             & \multicolumn{1}{c|}{Validation}       & Test             \\ \hline \hline
Ground-Truth                                 & \multicolumn{1}{c|}{$0.652 \pm 0.01$} & \multicolumn{1}{c||}{$0.623 \pm 0.02$} & \multicolumn{1}{c|}{$0.501 \pm 0.21$} & \multicolumn{1}{c||}{$0.489 \pm 0.16$} & \multicolumn{1}{c|}{$0.758 \pm 0.04$} & \multicolumn{1}{c||}{$0.701 \pm 0.09$} & \multicolumn{1}{c|}{$0.689 \pm 0.09$} & $0.663 \pm 0.05$ \\ \hline \hline
$\mathrm{CLBN}$  \cite{chow1968approximating}                           & \multicolumn{1}{c|}{$0.359 \pm 0.12$} & \multicolumn{1}{c||}{$0.317 \pm 0.09$} & \multicolumn{1}{c|}{$0.356 \pm 0.01$} & \multicolumn{1}{c||}{$0.318 \pm 0.02$} & \multicolumn{1}{c|}{$0.426 \pm 0.03$} & \multicolumn{1}{c||}{$0.411 \pm 0.05$} & \multicolumn{1}{c|}{$0.516 \pm 0.17$} & $0.510 \pm 0.22$ \\ \hline
$\mathrm{PrivBN}$ \cite{zhang2017privbayes}                      & \multicolumn{1}{c|}{$0.499 \pm 0.05$} & \multicolumn{1}{c||}{$0.436 \pm 0.14$} & \multicolumn{1}{c|}{$0.219 \pm 0.07$} & \multicolumn{1}{c||}{$0.202 \pm 0.01$} & \multicolumn{1}{c|}{$0.267 \pm 0.11$} & \multicolumn{1}{c||}{$0.267 \pm 0.07$} & \multicolumn{1}{c|}{$0.508 \pm 0.12$} & $0.479 \pm 0.08$ \\ \hline
$\mathrm{MedGAN}$  \cite{choi2017generating}                        & \multicolumn{1}{c|}{$0.367 \pm 0.08$} & \multicolumn{1}{c||}{$0.326 \pm 0.01$} & \multicolumn{1}{c|}{$0.183 \pm 0.10$} & \multicolumn{1}{c||}{$0.167 \pm 0.04$} & \multicolumn{1}{c|}{$0.314 \pm 0.06$} & \multicolumn{1}{c||}{$0.302 \pm 0.15$} & \multicolumn{1}{c|}{$0.534 \pm 0.01$} & $0.511 \pm 0.07$ \\ \hline
$\mathrm{TableGAN}$   \cite{park2018data}                    & \multicolumn{1}{c|}{$0.578 \pm 0.03$} & \multicolumn{1}{c||}{$0.509 \pm 0.06$} & \multicolumn{1}{c|}{$0.397 \pm 0.08$} & \multicolumn{1}{c||}{$0.382 \pm 0.17$} & \multicolumn{1}{c|}{$0.451 \pm 0.02$} & \multicolumn{1}{c||}{$0.427 \pm 0.09$} & \multicolumn{1}{c|}{$0.569 \pm 0.03$} & $0.557 \pm 0.06$ \\ \hline
$\mathrm{MotGAN}$   \cite{mottini2018airline}                       & \multicolumn{1}{c|}{$0.501 \pm 0.11$} & \multicolumn{1}{c||}{$0.487 \pm 0.07$} & \multicolumn{1}{c|}{$0.343 \pm 0.03$} & \multicolumn{1}{c||}{$0.316 \pm 0.02$} & \multicolumn{1}{c|}{$0.436 \pm 0.14$} & \multicolumn{1}{c||}{$0.405 \pm 0.19$} & \multicolumn{1}{c|}{$0.561 \pm 0.11$} & $0.523 \pm 0.14$ \\ \hline
$\mathrm{VeeGAN}$    \cite{srivastava2017veegan}                    & \multicolumn{1}{c|}{$0.588 \pm 0.14$} & \multicolumn{1}{c||}{$0.571 \pm 0.03$} & \multicolumn{1}{c|}{$0.410 \pm 0.05$} & \multicolumn{1}{c||}{$0.386 \pm 0.09$} & \multicolumn{1}{c|}{$0.459 \pm 0.07$} & \multicolumn{1}{c||}{$0.414 \pm 0.01$} & \multicolumn{1}{c|}{$0.564 \pm 0.06$} & $0.509 \pm 0.03$ \\ \hline
$\mathrm{ItsGAN}$ \cite{chen2019faketables}                       & \multicolumn{1}{c|}{$0.497 \pm 0.04$} & \multicolumn{1}{c||}{$0.448 \pm 0.01$} & \multicolumn{1}{c|}{$0.422 \pm 0.01$} & \multicolumn{1}{c||}{$0.392 \pm 0.02$} & \multicolumn{1}{c|}{$0.402 \pm 0.09$} & \multicolumn{1}{c||}{$0.384 \pm 0.10$} & \multicolumn{1}{c|}{$0.533 \pm 0.11$} & $0.501 \pm 0.12$ \\ \hline
$\mathrm{CctGAN}$   \cite{xu2019modeling}                        & \multicolumn{1}{c|}{$0.599 \pm 0.02$} & \multicolumn{1}{c||}{$0.581 \pm 0.08$} & \multicolumn{1}{c|}{$0.431 \pm 0.07$} & \multicolumn{1}{c||}{$0.403 \pm 0.01$} & \multicolumn{1}{c|}{$0.475 \pm 0.03$} & \multicolumn{1}{c||}{$0.471 \pm 0.02$} & \multicolumn{1}{c|}{$0.588 \pm 0.04$} & $0.572 \pm 0.01$ \\ \hline
$\mathrm{InvGAN}$  \cite{lee2021invertible}                         & \multicolumn{1}{c|}{$0.615 \pm 0.09$} & \multicolumn{1}{c||}{$0.601 \pm 0.05$} & \multicolumn{1}{c|}{$0.449 \pm 0.11$} & \multicolumn{1}{c||}{$0.432 \pm 0.06$} & \multicolumn{1}{c|}{$0.516 \pm 0.08$} & \multicolumn{1}{c||}{$0.494 \pm 0.05$} & \multicolumn{1}{c|}{$\mathbf{0.634 \pm 0.04}$} & $\mathbf{0.628 \pm 0.06}$ \\ \hline
$\mathrm{BctGAN}$  \cite{esmaeilpour2021bi}                         & \multicolumn{1}{c|}{$0.629 \pm 0.05$} & \multicolumn{1}{c||}{$0.611 \pm 0.01$} & \multicolumn{1}{c|}{$0.456 \pm 0.13$} & \multicolumn{1}{c||}{$0.424 \pm 0.10$} & \multicolumn{1}{c|}{$0.512 \pm 0.11$} & \multicolumn{1}{c||}{$0.508 \pm 0.09$} & \multicolumn{1}{c|}{$0.632 \pm 0.01$} & $0.626 \pm 0.03$ \\ \hline
$\mathrm{RccGAN}$    (ours)              & \multicolumn{1}{c|}{$\mathbf{0.641 \pm 0.02}$} & \multicolumn{1}{c||}{$\mathbf{0.629 \pm 0.05}$} & \multicolumn{1}{c|}{$\mathbf{0.478 \pm 0.01}$} & \multicolumn{1}{c||}{$\mathbf{0.475 \pm 0.03}$} & \multicolumn{1}{c|}{$\mathbf{0.602 \pm 0.04}$} & \multicolumn{1}{c||}{$\mathbf{0.598 \pm 0.01}$} & \multicolumn{1}{c|}{$0.617 \pm 0.07$} & $0.602 \pm 0.01$ \\ \hline
\end{tabular}}
\label{table:f1scoreClass}
\end{table*}

\subsection{Architectures for the Generator and Discriminator}
Unfortunately, there is no analytical approach for designing optimal architectures neither for the generator nor the discriminator in a GAN setup. Therefore, we empirically design such networks to accurately fit in the GAN configuration as depicted in Fig.~\ref{overview-GANarchitt}. The total number of hidden layers in both $\mathcal{G}$ and $\mathcal{D}$ should be manually tuned to make a balance between the number of required training parameters and the accuracy of the entire generative model.

Fig.~\ref{RccGAN-gen} illustrates the proposed architecture for the residual block of the $\mathrm{RccGAN}$. As shown, there are seven convolution layers with static receptive fields of $3\times 3$ and the stride of one. For simplicity purposes, we constantly set the number of filters to 128 for all these layers. Additionally, there are three batch normalization blocks ($\mathrm{BN}$) \cite{ioffe2015batch} either before or after a convolution layer. Technically, $\mathrm{BN}$ has two significant hyperparameters which effectively interferes with the system performance in runtime, namely the momentum and $\epsilon_{s}$ arguments where the latters denotes the stability term. For the majority of python frameworks (e.g., Pytorch), the momentum is initialized to 0.1 and $\epsilon_{s}=1e-5$. However, we slightly change these hyperparameters to enhance the performance of our $\mathrm{RccGAN}$. More specifically, we set the momentum argument to 0.22 and $\epsilon_{s}=1e-2$, upon conducting various exploratory experiments.

For adjusting the dimension of the output vectors for every block in the residual network configuration and counteracting with the potential noises on the candidate $\left \langle {\bar{\mu}}_{i},\bar{\sigma}_{i} \right \rangle$, two upsampling and one downsampling blocks with ratio 0.5 are embedded into Fig.~\ref{RccGAN-gen}. It is worth mentioning that we had an additional $\downarrow \mathrm{SM}$ block in our initial design but we eventually removed it since we noticed that it downshifts the skewness of $p_{c_{i,j}}$ (see Eq.~\ref{eq:pciandj}) during training.

Inspired from Xu {\it et al.}~\cite{xu2019modeling}, we also employ the Gumbel Softmax \cite{jang2016categorical} for the architecture of the residual network. In fact, we set the ratio of such an activation function to 0.16 for both blocks. According to our conducted experiments on the benchmarking databases, the confidence interval for the aforementioned ratio is $\left [0.09, 0.25 \right ]$. Ratios beyond this interval negatively affect the precision of the final output vector.

\begin{table*}[th]
\centering
\caption{Comparison of averaged $R^{2}$, $\mathrm{MSE}$, and $\mathrm{MAE}$ scores for the front-end regression algorithms trained on the synthesized tables using different generative models relative to the ground-truth (original database). We set the ratio of the validation and test subsets to 0.7 and 0.3, respectively. Outperforming results are shown in bold-face.}
\scriptsize
\resizebox{\textwidth}{!}{\begin{tabular}{r||cccccccc}
\hline
\multicolumn{1}{c||}{\multirow{4}{*}{Generative Model}} & \multicolumn{8}{c}{Tabular Database for the regression tasks}                                                                                                                                                                                                                                                                       \\ \cline{2-9} 
\multicolumn{1}{c||}{}                        & \multicolumn{4}{c||}{News}                                                                                                                                       & \multicolumn{4}{c}{King}                                                                                                               \\  \cline{2-9} 
\multicolumn{1}{c||}{}                        & \multicolumn{2}{c|}{$R^{2}$}                                                    & \multicolumn{2}{c||}{Error}                                                    & \multicolumn{2}{c|}{$R^{2}$}                                                  & \multicolumn{2}{c}{Error}                                \\ \cline{2-9} 
\multicolumn{1}{c||}{}                        & \multicolumn{1}{c|}{Validation}        & \multicolumn{1}{c|}{Test}              & \multicolumn{1}{c|}{$\mathrm{MSE}$}              & \multicolumn{1}{c||}{$\mathrm{MAE}$}              & \multicolumn{1}{c|}{Validation}       & \multicolumn{1}{c|}{Test}             & \multicolumn{1}{c|}{$\mathrm{MSE}$}              & $\mathrm{MAE}$              \\ \hline \hline
Ground-Truth                                 & \multicolumn{1}{c|}{$0.161 \pm 0.01$}  & \multicolumn{1}{c|}{$0.158 \pm 0.02$}  & \multicolumn{1}{c|}{$0.632 \pm 0.05$} & \multicolumn{1}{c||}{$0.587 \pm 0.09$} & \multicolumn{1}{c|}{$0.583 \pm 0.02$} & \multicolumn{1}{c|}{$0.544 \pm 0.10$} & \multicolumn{1}{c|}{$0.127 \pm 0.01$} & $0.331 \pm 0.02$ \\ \hline \hline
$\mathrm{CLBN}$ \cite{chow1968approximating}                            & \multicolumn{1}{c|}{$-4.124 \pm 0.12$} & \multicolumn{1}{c|}{$-4.560 \pm 0.04$} & \multicolumn{1}{c|}{$4.512 \pm 0.11$} & \multicolumn{1}{c||}{$6.027 \pm 0.02$} & \multicolumn{1}{c|}{$0.124 \pm 0.10$} & \multicolumn{1}{c|}{$0.119 \pm 0.13$} & \multicolumn{1}{c|}{$5.016 \pm 0.01$} & $7.446 \pm 0.19$ \\ \hline
$\mathrm{PrivBN}$ \cite{zhang2017privbayes}                             & \multicolumn{1}{c|}{$-3.045 \pm 0.23$} & \multicolumn{1}{c|}{$-3.193 \pm 0.07$} & \multicolumn{1}{c|}{$4.402 \pm 0.02$} & \multicolumn{1}{c||}{$5.805 \pm 0.09$} & \multicolumn{1}{c|}{$0.139 \pm 0.04$} & \multicolumn{1}{c|}{$0.131 \pm 0.07$} & \multicolumn{1}{c|}{$4.870 \pm 0.05$} & $7.005 \pm 0.09$ \\ \hline
$\mathrm{MedGAN}$  \cite{choi2017generating}                            & \multicolumn{1}{c|}{$-4.267 \pm 0.09$} & \multicolumn{1}{c|}{$-4.992 \pm 0.15$} & \multicolumn{1}{c|}{$5.639 \pm 0.04$} & \multicolumn{1}{c||}{$7.193 \pm 0.13$} & \multicolumn{1}{c|}{$0.223 \pm 0.05$} & \multicolumn{1}{c|}{$0.201 \pm 0.11$} & \multicolumn{1}{c|}{$4.725 \pm 0.09$} & $6.806 \pm 0.03$ \\ \hline
$\mathrm{TableGAN}$   \cite{park2018data}                         & \multicolumn{1}{c|}{$-2.134 \pm 0.08$} & \multicolumn{1}{c|}{$-2.886 \pm 0.09$} & \multicolumn{1}{c|}{$3.546 \pm 0.16$} & \multicolumn{1}{c||}{$4.255 \pm 0.01$} & \multicolumn{1}{c|}{$0.279 \pm 0.08$} & \multicolumn{1}{c|}{$0.234 \pm 0.02$} & \multicolumn{1}{c|}{$4.509 \pm 0.14$} & $6.566 \pm 0.08$ \\ \hline
$\mathrm{MotGAN}$  \cite{mottini2018airline}                            & \multicolumn{1}{c|}{$-5.671 \pm 0.15$} & \multicolumn{1}{c|}{$-6.044 \pm 0.01$} & \multicolumn{1}{c|}{$8.639 \pm 0.10$} & \multicolumn{1}{c||}{$8.776 \pm 0.06$} & \multicolumn{1}{c|}{$0.262 \pm 0.03$} & \multicolumn{1}{c|}{$0.210 \pm 0.04$} & \multicolumn{1}{c|}{$4.563 \pm 0.03$} & $6.592 \pm 0.02$ \\ \hline
$\mathrm{VeeGAN}$   \cite{srivastava2017veegan}                          & \multicolumn{1}{c|}{$-3.021 \pm 0.32$} & \multicolumn{1}{c|}{$-3.076 \pm 0.05$} & \multicolumn{1}{c|}{$4.119 \pm 0.02$} & \multicolumn{1}{c||}{$5.319 \pm 0.12$} & \multicolumn{1}{c|}{$0.280 \pm 0.06$} & \multicolumn{1}{c|}{$0.254 \pm 0.09$} & \multicolumn{1}{c|}{$4.301 \pm 0.01$} & $6.012 \pm 0.05$ \\ \hline
$\mathrm{ItsGAN}$    \cite{chen2019faketables}                          & \multicolumn{1}{c|}{$-2.414 \pm 0.07$} & \multicolumn{1}{c|}{$-3.128 \pm 0.17$} & \multicolumn{1}{c|}{$4.358 \pm 0.15$} & \multicolumn{1}{c||}{$5.771 \pm 0.03$} & \multicolumn{1}{c|}{$0.303 \pm 0.16$} & \multicolumn{1}{c|}{$0.292 \pm 0.08$} & \multicolumn{1}{c|}{$3.881 \pm 0.11$} & $5.310 \pm 0.02$ \\ \hline
$\mathrm{CctGAN}$    \cite{xu2019modeling}                          & \multicolumn{1}{c|}{$0.046 \pm 0.01$}  & \multicolumn{1}{c|}{$0.039 \pm 0.02$}  & \multicolumn{1}{c|}{$1.016 \pm 0.11$} & \multicolumn{1}{c||}{$2.129 \pm 0.14$} & \multicolumn{1}{c|}{$0.335 \pm 0.07$} & \multicolumn{1}{c|}{$0.299 \pm 0.02$} & \multicolumn{1}{c|}{$3.866 \pm 0.06$} & $5.323 \pm 0.12$ \\ \hline
$\mathrm{InvGAN}$   \cite{lee2021invertible}                           & \multicolumn{1}{c|}{$0.087 \pm 0.04$}  & \multicolumn{1}{c|}{$0.071 \pm 0.07$}  & \multicolumn{1}{c|}{$1.098 \pm 0.08$} & \multicolumn{1}{c||}{$2.518 \pm 0.05$} & \multicolumn{1}{c|}{$0.489 \pm 0.01$} & \multicolumn{1}{c|}{$0.403 \pm 0.17$} & \multicolumn{1}{c|}{$3.207 \pm 0.01$} & $4.117 \pm 0.01$ \\ \hline
$\mathrm{BctGAN}$  \cite{esmaeilpour2021bi}                            & \multicolumn{1}{c|}{$0.084 \pm 0.02$}  & \multicolumn{1}{c|}{$0.069 \pm 0.02$}  & \multicolumn{1}{c|}{$1.235 \pm 0.14$} & \multicolumn{1}{c||}{$2.781 \pm 0.09$} & \multicolumn{1}{c|}{$0.476 \pm 0.03$} & \multicolumn{1}{c|}{$0.396 \pm 0.06$} & \multicolumn{1}{c|}{$3.622 \pm 0.12$} & $4.458 \pm 0.08$ \\ \hline
$\mathrm{RccGAN}$  (ours)                          & \multicolumn{1}{c|}{$\mathbf{0.124 \pm 0.03}$}  & \multicolumn{1}{c|}{$\mathbf{0.115 \pm 0.01}$}  & \multicolumn{1}{c|}{$\mathbf{0.976 \pm 0.05}$} & \multicolumn{1}{c||}{$\mathbf{1.393 \pm 0.03}$} & \multicolumn{1}{c|}{$\mathbf{0.485 \pm 0.04}$} & \multicolumn{1}{c|}{$\mathbf{0.422 \pm 0.01}$} & \multicolumn{1}{c|}{$\mathbf{2.001 \pm 0.07}$} & $\mathbf{2.065 \pm 0.03}$ \\ \hline
\end{tabular}
}
\label{table:R2ScoreReg}
\end{table*}

The last block in Fig.~\ref{RccGAN-gen} is the concatenation operator which appends the output of the last convolution layer with the Gumbel softmax vector. Technically, this operation forces the entire residual block to avoid memorizing $\left \langle {\bar{\mu}}_{i},\bar{\sigma}_{i} \right \rangle$ derived from the convolution layer and consequently simplifies the task of the discriminator network. In other words, the output vector of $\mathrm{Gum.}$ resembles a label for the $\mathrm{Conv.}$ layer and it helps to discriminate synthetic records with close mean and variance parameters. Therefore, no complex architecture is required for the discriminator and this considerably improves the computational complexity of the generative model. 

Our proposed architecture for $\mathcal{D}(\cdot)$ is a three-layered convolutional neural network which receives the synthetic records from the generator and validates their similarity with $p_{r}$ in each batch with size of 512. The number of filters in the first two layers is constantly 64 but the last layer incorporates 128 filters followed by a $\mathrm{BN}$ and ReLU activation function. For training, we use Adam optimizer \cite{kingma2014adam} with $\left [ 0, 0.85 \right ]$ parameters and $1.2\cdot 10^{-4}$ learning rate on ten NVIDIA GTX-1080-Ti GPU with $12\times 11$ memory. In the following subsection, we provide the results achieved from training this generative model on the experimental databases mentioned in Section~\ref{sec:databasesub}.

\subsection{Achieved Results}
We train $\mathrm{RccGAN}$ and the baseline generative models as explained in Section~\ref{sec:background} on the six benchmarking tabular databases (main tables). Therefore, for every GAN associated with an original database, a single table will be synthesized. Then, we randomly select 70\% of the entire records in both $\mathcal{T}_{org}$ and $\mathcal{T}_{syn}$ for separately training some front-end classification and regression algorithms. Toward this end, we mostly follow the experimental protocol suggested in \cite{xu2019modeling, esmaeilpour2021bi}. More specifically, for Adult, Census, HCDR, and Cabs databases, we implement decision tree (with depth of 20 to 30), convolutional neural network (with 10 to 15 hidden layers), Adaboost (with 10 to 50 estimators), and ResNet (with 18 and 34 hidden layers). Additionally, we fit the nonlinear regression algorithm with polynomial kernels (with degree two to five) and deep neural network (with 20 to 40 hidden layers). For fairness in evaluation, we run the five-fold cross validation for every algorithm and select the most comprehensive model for testing on the remaining 30\% portion of the original and synthesized tables. Finally, we compute either the averaged $F_{1}$ (in-place-of classification) or $R^{2}$ (in-place-of regression) for the achieved models and compare the scores with their associated ground-truth tables.

Table~\ref{table:f1scoreClass} presents our achieved results associated with the tabular databases which have been collected for the classification tasks, namely Adult, Census, HCDR, and Cabs. As shown, for the majority of the cases, the ensemble of the aforementioned classifiers with identical settings and configurations achieve higher $F_{1}$ scores on the datasets synthesized by $\mathrm{RccGAN}$. In other words, our proposed generative model outperforms others in synthesizing tables with higher similarity relative to the ground-truth. However, it competitively fails against both $\mathrm{InvGAN}$ and $\mathrm{BctGAN}$ for the Cabs database. Our conjecture is that multiple residual layers embedded in the $\mathrm{RccGAN}$ unnecessarily increases the number of parameters to train on this small database and this negatively affects the generalizability of our entire model. We investigated this issue through removing the middle residual block and $\mathrm{Skip}_{L \rightarrow R}$ in Fig.~\ref{overview-GANarchitt}. Such a modification increased the $F_{1}$ score to $0.657 \pm 0.02$ and $0.643 \pm 0.06$ for validation and test subsets, respectively.         

Table~\ref{table:R2ScoreReg} summarizes our achieved results on the tabular databases assembled for the regression tasks, namely News and King. We compute $R^{2}$ scores for the front-end regression algorithms which have been trained on the tabular databases synthesized by various generative models. Inspired by \cite{lee2021invertible}, we also measure the regression error using mean squared error ($\mathrm{MSE}$) and mean absolute error ($\mathrm{MAE}$) metrics. Comparing such scores with the ground-truth values corroborates that our $\mathrm{RccGAN}$ outperforms other generative models. For evaluating the scalability of our proposed GAN, we repeated the cross validation operation with various folds from three to ten and we achieved almost the same performance.

Regarding Table~\ref{table:R2ScoreReg}, there are some generative models with negative $R^{2}$ which indicate that their associated front-end regression models could not accurately draw the decision boundary. However, it is worth mentioning that even $R^{2}$ score for the ground-truth table is not satisfactorily high. In the following subsection, we analyze the stability of all these generative models during training and identify their potential pitfalls. 

\begin{figure*}[th]
  \centering
  \includegraphics[width=\textwidth]{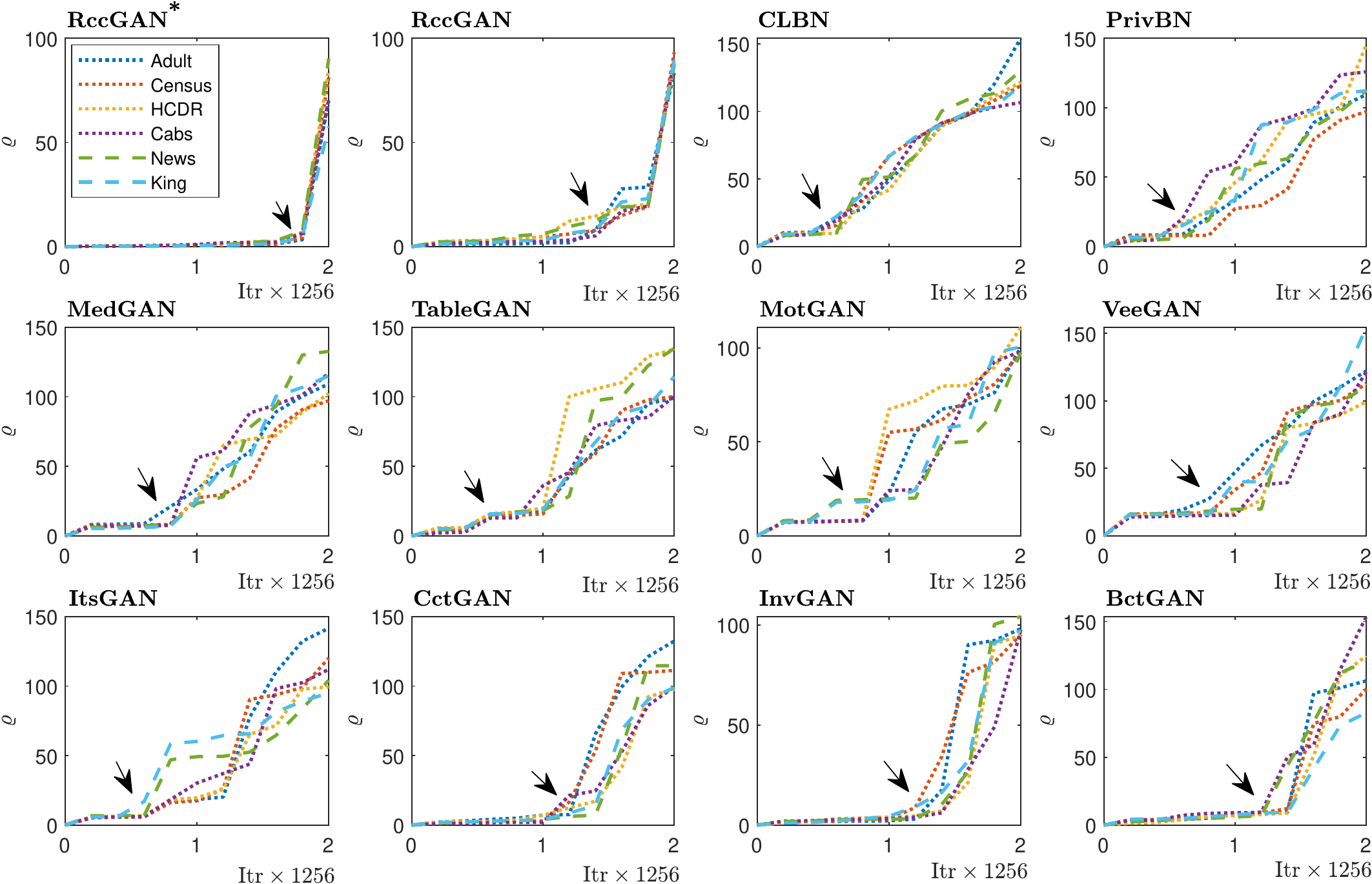}
  \caption{Monitoring the stability of the tabular GANs during training on six benchmarking databases. Curves associated with the classification and regression tasks are depicted by dotted and dashed formats, respectively. Values along $x$-axis denote the average of singular values for $\theta_{g}$ and those along $y$-axis indicate the training iterations. Herein, the onset of instability (explosion) for every model is pointed out by a black arrow. The difference between $\mathrm{RccGAN^{*}}$ and $\mathrm{RccGAN}$ is that the latter does not employ the proposed regularization technique as explained in Section~\ref{sec:subReguProtc}.}  
  \label{fig:stabilityMonit}
\end{figure*}

\subsection{Stability Monitoring}
Amongst the most straightforward approaches developed for monitoring the stability of the GANs is plotting the distribution of the weight vectors associated with the generator and/or discriminator networks during training (i.e., $\theta_{g}$ and $\theta_{d}$, respectively) \cite{odena2018generator}. This contributes to identify the onset of potential anomalies (e.g., explosion or sudden magnitude elevation) for such vectors. However, the dimensionality of both $\theta_{g}$ and $\theta_{d}$ is relatively high and plotting their distributions which usually carry numerous tiny fluctuations might result in misinterpretations. For addressing this concern, Brock {\it et al.}~\cite{brock2018large} suggested to plot the distribution of some singular values associated to these vectors. More specifically, they factorize $\theta_{g}$ into three matrices using the conventional singular value decomposition (SVD) algorithm as \cite{van1983matrix}: 
\begin{equation}
    \theta_{g} \mapsto \theta_{h} \times \theta_{s} \times \theta_{a}^{\top}	
\end{equation}
\noindent where $\theta_{h}$, $\theta_{a}$, and $\theta_{s}$ denote the hanger, aligner, and stretcher matrices derived from such a decomposition operation. Commonly, the first two matrices are known as the container of left and right singular vectors and the latter matrix includes the singular values. It is worth mentioning that $\theta_{s}$ is a sparse matrix and except for the diagonal entries (where the singular values reside) all other fields are zero. In other words, $\theta_{s}$ can be defined as the following \cite{van1983matrix}:  
\begin{equation}
    \theta_{s} [i,j] = \left\{\begin{matrix}
\varrho_{i} & \mathrm{if} \quad i=j \\ 
0 & \mathrm{if} \quad i\neq j 
\end{matrix}\right.
\end{equation}
\noindent where $i$ and $j$ represent the row and columns indices, respectively. Additionally, $\varrho_{i}$ denotes the $i$-th singular value and:
\begin{equation}
    \mathrm{diag}\left ( \theta_{s} \right )=\left [ \varrho_{0}, \varrho_{1},\cdots \varrho_{i} \right ]_{i=1}^{m_{r}}
    \label{eq:diagthetaS}
\end{equation}
\noindent where $\mathrm{diag}(\cdot)$ returns the diagonal entries of the given matrix and $m_{r}$ refer to the total number of rows in $\theta_{s}$.  

Mathematically, one of the most significant characteristics of Eq.~\ref{eq:diagthetaS} is that values in the vector of $\mathrm{diag}\left ( \theta_{s} \right )$ are sorted in a descending order (in terms of magnitude) from the most valuable component (i.e., $\varrho_{0}$) to the least informative coefficient (i.e., $\varrho_{m_{r}}$) \cite{baker2005singular}. Based on this pivotal algebraic axiom, Brock {\it et al.}~\cite{brock2018large} exploit only the top three singular values (i.e., $\varrho_{0}$, $\varrho_{1}$, and $\varrho_{2}$) for monitoring the distribution of $\theta_{g}$ during training. However, they analyze such values separately and without incorporating the correlation among them. Unfortunately, this might negatively affect the procedure of accurately tracking critical changes in $\theta_{g}$. For addressing this concern, we introduce a new stability monitoring metric as follows:
\begin{equation}
    \varrho:=\underbrace{\varrho_{i_{c} \rightarrow 0}}_{\text{anchor}}+\underbrace{\sum_{i_{c}=0}^{m_{\hat{r}}}\frac{\varrho_{i_{c}}}{\varrho_{i_{c}+1}+1} + b_{0}}_{\text{partial-correlation}}
    \label{eq:corrrvarho}
\end{equation}
\noindent where the anchor term denotes the premier singular value \cite{van1976generalizing} for the weight vectors and if a sudden elevation (or any perturbation) occurs for $\theta_{g}$, it will directly reflect on $\varrho_{o}$ \cite{van1983matrix}. Hence, incorporating such an anchor value is critical for monitoring $\theta_{g}$. The second term in Eq.~\ref{eq:corrrvarho} relates to computing the partial-correlation ($\mathrm{PC}$) for singular values \cite{baba2004partial}. Statistically, it approximates the relative divergence among the components of $\mathrm{diag}\left ( \theta_{s} \right )$ and yields a nonlinear function for extracting its local dependencies \cite{kendall1948advanced}. The main intuition behind employing $\mathrm{PC}$ over the conventional metrics such as Pearson \cite{benesty2009pearson}, Kendall \cite{abdi2007kendall}, and Spearman correlation coefficients \cite{myers2004spearman} is its extreme simplicity and efficiency at runtime. 

As shown in Eq.~\ref{eq:corrrvarho}, $\mathrm{PC}$ enquires two hyperparameters, namely $m_{\hat{r}}$ and $b_{0}$. The first hyperparameter refers to the total number of singular values which should be included for accurately approximating $\mathrm{PC}$. Assigning large values to $m_{\hat{r}}$ dramatically increases the computational complexity for the SVD algorithm. Therefore, we empirically set:
\begin{equation}
m_{\hat{r}} = c_{s} \cdot m_{r} \quad \mathrm{for} \quad     c_{s}\in\left [ 0.05, 0.15 \right ]
\end{equation}
\noindent in order to make a reasonable trade-off between the precision of $\mathrm{PC}$ and effectively reducing the computational overhead for the SVD algorithm. Allocating values beyond the aforementioned confidence interval for $c_{s}$ might compromise such a trade-off.

\begin{table*}[th]
\centering
\caption{Comparing the reliability scores of the proposed $\mathrm{RccGAN}$ for different tabular databases with two scales. Herein, the scale of 0.5 denotes the 50\% of the table's records which they have been randomly selected and the scale of 1 refers to the entire table. All the reliability scores are averaged over ten times repetition of experiments following the ANOVA test \cite{cuevas2004anova}.}
\scriptsize
\resizebox{\textwidth}{!}{\begin{tabular}{c||cccccccccccc}
\hline
\multirow{3}{*}{\begin{tabular}[c]{@{}c@{}}Reliability\\ Metric\end{tabular}} & \multicolumn{12}{c}{Tabular databases}                                                                                                                                                                                                                                                                                                                                                                                                                                                                       \\ \cline{2-13} 
                                                                              & \multicolumn{2}{c||}{Adult}                                                          & \multicolumn{2}{c||}{Census}                                                         & \multicolumn{2}{c||}{HCDR}                                                           & \multicolumn{2}{c||}{Cabs}                                                           & \multicolumn{2}{c||}{News}                                                           & \multicolumn{2}{c}{King}                                       \\ \cline{2-13} 
                                                                              & \multicolumn{1}{c|}{$\mathrm{Scale}=0.5$} & \multicolumn{1}{c||}{$\mathrm{Scale}=1$} & \multicolumn{1}{c|}{$\mathrm{Scale}=0.5$} & \multicolumn{1}{c||}{$\mathrm{Scale}=1$} & \multicolumn{1}{c|}{$\mathrm{Scale}=0.5$} & \multicolumn{1}{c||}{$\mathrm{Scale}=1$} & \multicolumn{1}{c|}{$\mathrm{Scale}=0.5$} & \multicolumn{1}{c||}{$\mathrm{Scale}=1$} & \multicolumn{1}{c|}{$\mathrm{Scale}=0.5$} & \multicolumn{1}{c||}{$\mathrm{Scale}=1$} & \multicolumn{1}{c|}{$\mathrm{Scale}=0.5$} & $\mathrm{Scale}=1$ \\ \hline \hline
\multicolumn{1}{r||}{$\mathrm{NNDR}$ \cite{mendes2017nearest}}                                          & \multicolumn{1}{c|}{$0.914$}              & \multicolumn{1}{c||}{$0.901$}            & \multicolumn{1}{c|}{$0.856$}              & \multicolumn{1}{c||}{$0.823$}            & \multicolumn{1}{c|}{$0.941$}              & \multicolumn{1}{c||}{$0.905$}            & \multicolumn{1}{c|}{$0.873$}              & \multicolumn{1}{c||}{$0.829$}            & \multicolumn{1}{c|}{$0.956$}              & \multicolumn{1}{c||}{$0.910$}            & \multicolumn{1}{c|}{$0.965$}              & $0.919$            \\ \hline
\multicolumn{1}{r||}{$\mathrm{IMS}$ \cite{elmagarmid2006duplicate}}                                           & \multicolumn{1}{c|}{$0.000$}              & \multicolumn{1}{c||}{$0.000$}            & \multicolumn{1}{c|}{$0.000$}              & \multicolumn{1}{c||}{$0.000$}            & \multicolumn{1}{c|}{$0.000$}              & \multicolumn{1}{c||}{$0.000$}            & \multicolumn{1}{c|}{$0.000$}              & \multicolumn{1}{c||}{$0.000$}            & \multicolumn{1}{c|}{$0.000$}              & \multicolumn{1}{c||}{$0.000$}            & \multicolumn{1}{c|}{$0.000$}              & $0.000$            \\ \hline
\multicolumn{1}{r||}{$\mathrm{DCR}$ \cite{corral2000closest}}                                           & \multicolumn{1}{c|}{$0.834$}              & \multicolumn{1}{c||}{$0.825$}            & \multicolumn{1}{c|}{$0.769$}              & \multicolumn{1}{c||}{$0.739$}            & \multicolumn{1}{c|}{$0.681$}              & \multicolumn{1}{c||}{$0.665$}            & \multicolumn{1}{c|}{$0.979$}              & \multicolumn{1}{c||}{$0.953$}            & \multicolumn{1}{c|}{$0.803$}              & \multicolumn{1}{c||}{$0.787$}            & \multicolumn{1}{c|}{$0.715$}              & $0.706$            \\ \hline
\end{tabular}
}
\label{table:reliabilityconds}
\end{table*}

The second hyperparameter for computing $\mathrm{PC}$ is the bias term and it should be defined according to the distribution of $\theta_{g}$. The motivation behind embedding $b_{0}$ in Eq.~\ref{eq:corrrvarho} is balancing the anchor term when $\theta_{g}$ is semi-unitary, null, or circulant and running the SVD operation might result in ill-conditioned cases for $\mathrm{PC}$ \cite{horn2012matrix}. Thus, when there is no specific irregularities in $\theta_{g}$, we can initialize such a bias term to zero. 

We employ our proposed metric (Eq.~\ref{eq:corrrvarho}) for monitoring the stability of the generators trained on the six benchmarking tabular databases. More specifically, we plot the value of $\varrho$ over the number of iterations in order to identify any signs of instability in $\theta_{g}$ during training. The achieved results are illustrated in Fig.~\ref{fig:stabilityMonit}. As shown, for the majority of the cases, the early signs of instability emerges between 700 to 1.3K iterations. However, our proposed $\mathrm{RccGAN}$ delays such signs until 1.46k iterations and those results are obtained without using the regularization protocol (as explained in Section~\ref{sec:subReguProtc}). This corroborates the hypothesis that our proposed GAN is inherently capable to synthesize any large-scale tabular databases.

For measuring the absolute impact of exploiting our proposed regularization protocol (Eq.~\ref{eq:intervalRegul}) on the training performance of $\mathrm{RccGAN}$, we carefully track the distribution of $\varrho$ for $\theta_{g}$. As depicted in Fig.~\ref{fig:stabilityMonit}, such a regularization further delays the onset of instability until 1.93k iterations. More specifically, our proposed regularization technique results in 24.3\% improvement for the stability of the generator during training. According to our conducted experiments, employing Eq.~\ref{eq:intervalRegul} for other generative models enhances their stability approximately 3.4\% (on average) which is still below the generic $\mathrm{RccGAN}$ (without regularization).

Our major motivation to employ Eq.~\ref{eq:corrrvarho} only for monitoring the stability of the generator and skipping to report such an experiment for tracking $\theta_{d}$ is two fold. Firstly, instability often prohibits the performance of the generator network and it results in more destructive side-effects on $\mathcal{G}(\cdot)$. Secondly, we carried out several experiments to identify irregularities for the distribution of $\varrho$ in $\theta_{d}$ but we could not clearly spot the onset of potential instabilities. This might be interpreted that the discriminator networks are fairly more stable than the generator networks on our benchmarking tabular databases.

In the following subsection, we investigate the compliance of the generative models with the reliability conditions required for synthesizing a tabular database.

%clarkson1999nearest
\subsection{Investigating the Reliability Conditions for RccGAN}
In Section~\ref{sec:introduction}, we defined three reliability conditions for a generative model in the context of synthesizing comprehensive tabular databases. The first condition obliges to compute the non-parametric probability distribution \cite{kimball1947some} between $\mathcal{T}_{syn}$ and $\mathcal{T}_{org}$ (noteworthy, the generator network models the parametric distribution). Such a computation contributes to fairly compare potential discrepancies between two tables. Toward this end, we measure the nearest neighbor distance ratio ($\mathrm{NNDR}$) metric \cite{mendes2017nearest}. Technically, $\mathrm{NNDR}(\cdot)$ measures the cross-correlation between two probability density functions (i.e., $\mathbb{P}(\cdot)$ and $\mathbb{Q}(\cdot)$ respectively for the original and synthesized tables) and it is in close relation with the inverse Jensen–Shannon divergence \cite{clarkson1999nearest,fuglede2004jensen} in such a way that:
\begin{equation}
   0\leq \mathrm{NNDR}\left [ \mathbb{P}(\mathcal{T}_{org}),\mathbb{Q}(\mathcal{T}_{syn}) \right ] \leq 1. 
   \label{eq:nndr}
\end{equation}
\noindent This implies that when $\mathbb{P}(\mathcal{T}_{org})$ and $\mathbb{Q}(\mathcal{T}_{syn})$ are fully divergent, their $\mathrm{NNDR}$ score is zero.

The second reliability condition enquires to verify that whether the synthetic records are unanimously distinguishable from the originals. In other words, the number of identical records in $\mathcal{T}_{org}$ and $\mathcal{T}_{syn}$ should be zero to comply with such a condition. Toward searching for potential duplicated records, we employ the identical match share ($\mathrm{IMS}$) metric \cite{elmagarmid2006duplicate}. Technically, $\mathrm{IMS}$ compares every record in the synthesized dataset with its counterpart in the original table. Therefore, according to these explanations, the ideal case for $\mathrm{IMS}$ is being constantly zero. 

Finally, the third reliability condition is tracing the possibility of identifying original records from $\mathcal{T}_{syn}$. To the best of our knowledge, there is no specific metric developed for such a task. However, we utilize the distance to the closest record ratio ($\mathrm{DCR}$) metric \cite{corral2000closest}. From a statistical and algebraic point of view, $\mathrm{DCR}$ measures the adjacency of manifolds associated with pairs of original and synthetic records. Unfortunately, there is no upperbound for such a metric but obtaining larger values for $\mathrm{DCR}$ interprets as lower chance of tracing an original record from its synthetic counterpart. In order to make the $\mathrm{DCR}$ scores more comparable, we normalize all the achieved values between zero and one. Thus, yielding a score closer to one indicates a higher reliability of the generative model.

Table~\ref{table:reliabilityconds} summarizes the achieved scores for the three above-mentioned reliability metrics associated with our proposed $\mathrm{RccGAN}$. As shown in this table, we have reported all the scores for two scales of 0.5 and 1. More specifically, we separately compute $\mathrm{NNDR}$, $\mathrm{IMS}$, and $\mathrm{DCR}$ values once for half of the tables and then for the entire database. Our main motivation for running such experiments is that we noticed the reliability of the GANs varies from small to large scales. Our investigation corroborates that the reliability of our GAN slightly drops when the size of $\mathcal{T}_{org}$ increases. The ratio of such a drop is considerably higher (i.e., 0.08 on average) for other generative models as discussed in Section~\ref{sec:background}. Following these results, we repeated our experiments for Table~\ref{table:f1scoreClass}, Table~\ref{table:R2ScoreReg}, and Fig.~\ref{fig:stabilityMonit} on different scales for $\mathcal{T}_{org}$. Our achieved results confirmed no tangible difference between various scales from 0.5 to 1.

In the following Section, we provide additional discussions on different aspects of our proposed tabular synthesis approach and the conducted experiments.

%%%%%%%%%%%%%%%%%%%%%%%%%%%%%%%%%%%%%%%%%%%%%%%%%%%%%%%%%

%%%%%%%%%%%%%%%%%%%%%%%%%%%%%%%%%%%%%%%%%%%%%%%%%%%%%%%%%
\section{Discussions}
\label{sec:discussions}
This section addresses some discussions on both the theory and implementation of the $\mathrm{RccGAN}$. Additionally, we extend the application of our proposed regularization technique to the GANs on other data modalities such as environmental audio and speech.

\subsection{Precision of the Designed Conditional Vector}
In practice, there are two potential flaws in employing $\mathcal{CD}_{d}$ and $\mathcal{CD}_{b}$ for implementing Eq.~\ref{eq:condiGeni}. Firstly, the statistical mode is not explicitly defined for Cantor distribution \cite{morrison1998random} Therefore, conditioning the generator network to vectors drawn from such probability distributions (i.e., $\eta_{k_{d},d}$ and $\eta_{k_{b},b}$) might result in memorizing a few random modes of the training table, particularly when the majority of the $C_{i}$s are either binary or discrete. One potential resolution for addressing this issue is to randomly shuffling all the columns of $\mathcal{T}_{org}$ in every iteration during training $\mathcal{G}(\cdot)$. However, this might not be feasible for super large-scale tabular databases.

Secondly, concatenating $\eta_{k_{d},d}$ and $\eta_{k_{b},b}$ as defined in Eq.~\ref{eq:condiGeni} forces the generator to incorporate both the discrete and binary columns for synthesizing new records even if such $C_{i}$s contain null or empty values. In response to this concern, we should either replace all the null fields with random values before starting to train $\mathcal{G}(\cdot)$ or embedding a new term in Eq.~\ref{eq:condiGeni} for regularizing null values. Since the latter approach might not be computationally efficient in runtime, we opt to the first suggestion in our entire experiments.

\subsection{Replacing Residual Block with Fully Convolutional}
Our conducted experiments reveals that replacing the residual blocks in Fig.~\ref{overview-GANarchitt} with only convolutional layers (for the sake of simplifying the configuration of the GAN) dramatically degrades the performance of the $\mathrm{RccGAN}$. More specifically, such a replacement drops the $F_{1}$ and $R^{2}$ scores as reported in Table~\ref{table:f1scoreClass} and Table~\ref{table:R2ScoreReg} respectively around 41.2\% and 32.16\% averaged over the six experimental databases. Additionally, it advances the onset of instabilities about 45.82\% (without the regularization term in Eq.~\ref{eq:intervalRegul}) and 26.8\% (with regularization) compared to the original $\mathrm{RccGAN}$ as depicted in Fig.~\ref{fig:stabilityMonit} (the top left sub-figure). However, it is worth mentioning that, every residual block in Fig.~\ref{overview-GANarchitt} requires 19.8\% more training parameters than a convolutional block with the same number of filters.

\subsection{Multiple Discriminator Networks}
The major task of a discriminator in a GAN setup is providing gradients to the generator network during training \cite{goodfellow2016deep}. Therefore, employing multiple discriminators might increase the chance of returning more informative gradients to $\mathcal{G}(\cdot)$ and consequently improving the stability of the generator \cite{nguyen2017dual}. Toward this end, we investigate the stability of our proposed $\mathrm{RccGAN}$ with multiple discriminator networks. Our achieved results corroborate that it is possible to delay the onset of instabilities for extra 500 iterations compared to Fig.~\ref{fig:stabilityMonit} with three discriminators. However, this operation triples the number of required training parameters and prohibitively increases the computational complexity for the entire GAN.

%%%%%%%%%%%%%%%%%%%%%%%%%%%%%%%%%%%%%%%%%%%%%%%%%%%%%%%%%

\section{Conclusion}
\label{sec:conclusion}
%%%%%%%%%%%%%%%%%%%%%%%%%%%%%%%%%%%%%%%%%%%%%%%%%%%%%%%%%
In this paper, we introduced a novel generative adversarial network for synthesizing large-scale tabular databases. Unlike both the conventional and cutting-edge class-conditioned GANs which employ the long-tailed mask functions for deriving the conditional vector, we proposed a simpler formulation. In fact, we bypassed the implementation of the costly mask function by using two independent Cantor distributions associated with binary and discrete features. Such a formulation not only reduces the length of the conditional vector but also helps the generator network to better capturing the training samples distribution. With respect to the heterogeneous nature of the tabular databases, we designed a new GAN configurations using a stack of residual blocks. For improving the stability of the entire model we imposed a noble regularization protocol on the generator to more effectively control the variation of its weight vectors. According to the reported results, such an operation considerably delayed the onset of instabilities for our proposed $\mathrm{RccGAN}$. Furthermore, for closely monitoring the stability of the generator network during training, we developed a new metric for identifying irregularities or sudden perturbation on $\theta_{g}$. Finally, we verified the reliability of our $\mathrm{RccGAN}$ using NNDR, IMS, and DCR metrics. Our achieved results confirm that such a synthesis approach fully complies with the three conditions of the reliability concept. However, we noticed slight drop for NNDR and DCR metrics on databases with different scales. We are determined to address this issue in our future works.

\section*{Acknowledgment}
This work was funded by Fédération des Caisses Desjardins du Québec and Mitacs accelerate program with agreement number IT25105.

\bibliographystyle{IEEEtran}
\bibliography{IEEEabrv,mybib}

\vfill


\end{document}